\documentclass[runningheads]{llncs}

\usepackage{eccv}

\usepackage{eccvabbrv}

\usepackage{graphicx}
\usepackage{booktabs}

\usepackage{multirow}
\usepackage{breqn}
\usepackage{algorithm} 
\usepackage{algorithmic}
\usepackage{csquotes}
\usepackage{caption} 
\usepackage{subcaption}
\usepackage[accsupp]{axessibility}  % Improves PDF readability for those with disabilities.

\usepackage{hyperref}

\usepackage{orcidlink}

\begin{document}

% \title{BehAVE: Behaviour Alignment of Video Game Encodings}
\title{BehAVE: Behaviour Alignment of \\ Video Game Encodings}

\titlerunning{BehAVE}


\author{Nemanja Rašajski\inst{1}\thanks{Equal contribution}\orcidlink{0009-0000-3487-1339} \and
Chintan Trivedi\inst{1}*\orcidlink{0000-0003-2749-2618} \and
Konstantinos Makantasis\inst{2}\orcidlink{0000-0002-0889-2766} \and 
Antonios Liapis\inst{1}\orcidlink{0000-0001-5554-1961}
\and
Georgios N. Yannakakis\inst{1}\orcidlink{0000-0001-7793-1450}}

\authorrunning{Rašajski et al.}

\institute{Institute of Digital Games, University of Malta, Malta 
\email{\{nemanja.rasajski,ctriv01,antonios.liapis,georgios.yannakakis\}@um.edu.mt}\\
\and
AI Department, University of Malta, Malta\\
\email{konstantinos.makantasis@um.edu.mt}}

\maketitle

\begin{abstract}

Domain randomisation enhances the transferability of vision models across visually distinct domains with similar content. However, current methods heavily depend on intricate simulation engines, hampering feasibility and scalability. This paper introduces BehAVE\footnote{\url{https://sites.google.com/view/behavefw/home}}, a video understanding framework that utilises existing commercial video games for domain randomisation without accessing their simulation engines. BehAVE taps into the visual diversity of video games for randomisation and uses textual descriptions of player actions to \emph{align} videos with similar content. We evaluate BehAVE across 25 first-person shooter (FPS) games using various video and text foundation models, demonstrating its robustness in domain randomisation. BehAVE effectively aligns player behavioural patterns and achieves zero-shot transfer to multiple unseen FPS games when trained on just one game. In a more challenging scenario, BehAVE enhances the zero-shot transferability of foundation models to unseen FPS games, even when trained on a game of a different genre, with improvements of up to 22\%. BehAVE is available online\footnote{\url{https://github.com/nrasajski/BehAVE}}. 
\end{abstract}

% nrasajski/BehAVE
% anonymous/repo
    
\section{Introduction}
\label{sec:intro}

% \begin{figure}[!tb] 
%         \centering \includegraphics[height=0.6\linewidth, width=0.9\linewidth]{figures/behavior_tsne2.png}
%         \caption{Impact \textbf{BehAVE} framework. The t-SNE plots show encodings of short video sequences from 5 distinct FPS games: (a) indicates the \textit{domain gap} between encodings of different games from a video foundation model, while (b) shows encodings aligned by BehAVE. The framework positions similar player behavior encodings (\emph{e.g.}, aim gun) closely across visually diverse games like \textit{PUBG} (left) and \textit{Apex Legends} (right).}
%         \label{fig:alignmentdemo}
% \end{figure}

\begin{figure}[!tb]
    \centering
    \begin{subfigure}[b]{0.49\textwidth}
        \centering
        \includegraphics[width=\textwidth]{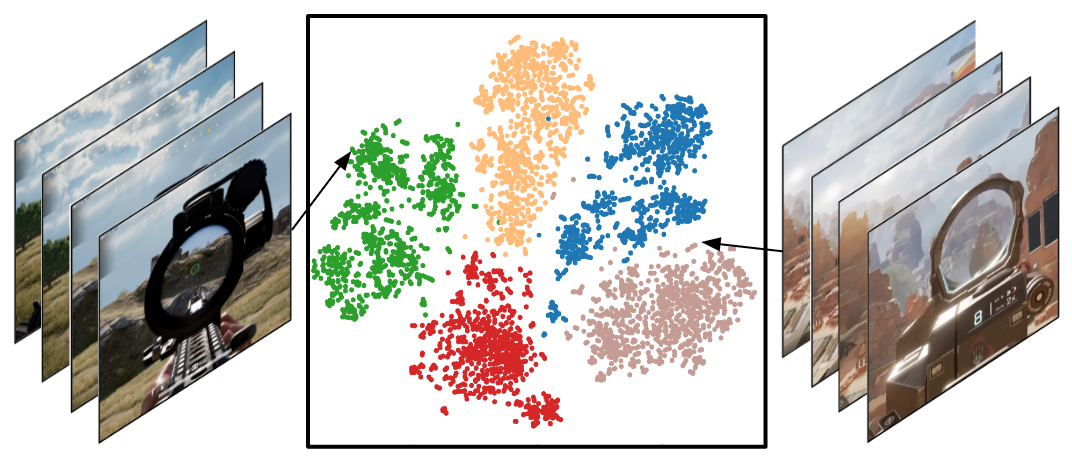}
        \caption{Foundation Encodings}
        \label{fig:demo_foundation}
    \end{subfigure}
    % \hfill
    \begin{subfigure}[b]{0.49\textwidth}
        \centering
        \includegraphics[width=\textwidth]{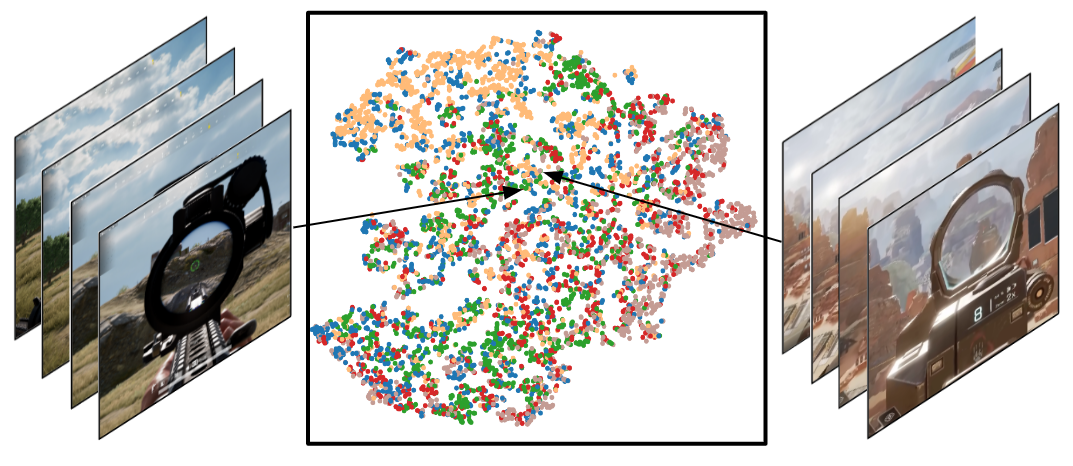}
        \caption{BehAVE Encodings}
        \label{fig:demo_behave}
    \end{subfigure}
    \caption{High level overview of the \textbf{BehAVE} framework. The t-SNE plots show encodings of short video sequences from 5 distinct FPS games: (a) indicates the \textit{domain gap} between encodings of different games from a video foundation model, while (b) shows encodings aligned by BehAVE. The framework positions similar player behaviour encodings (\emph{e.g.}, aim gun) closely across visually diverse games like \textit{PUBG} (left) and \textit{Apex Legends} (right).}
    \label{fig:alignmentdemo}
\end{figure}

Video game engines uphold an internal representation of the game environment \cite{lewis2002game, rabin2005introduction}, encompassing essential variables such as player position and map layout. Upon undergoing processing by the game graphics \emph{renderer}, this data becomes intricately entwined with the game's visual style, resulting in the images presented to the player on screen. Securing access to game engine data, however, proves challenging, if not impossible in practice, particularly for commercial video games. Consequently, this circumstance directs the trajectory of game artificial intelligence (AI) research towards the utilisation of more accessible game representations such as pixels \cite{mnih2013playing, wydmuch2018vizdoom}. Unfortunately even with state of the art pre-trained computer vision (CV) models the resulting game pixel encodings do not generalise well, even between games of the same genre, and suffer from what is known as the \emph{domain gap} problem \cite{wang2018deep} (see Fig. \ref{fig:demo_foundation}).

Improving model generalisation stands as a pivotal problem within machine learning (ML) research, encompassing various research topics such as domain adaptation, meta-learning, and transfer learning. Domain generalisation \cite{wang2022generalizing} has emerged as a focal point of interest aiming to develop models capable of effectively generalising across unseen testing domains by leveraging training data from diverse but related domains. One highly promising technique for enhancing the transferability of CV models in games is \textbf{domain randomisation} \cite{tobin2017domain}, a simple technique that improves the robustness of a CV model by training it on visuals derived from randomising rendering parameters of a simulator engine. Building accurate large-scale simulators, however, is a formidable challenge which requires significant time, expertise, and effort \cite{dosovitskiy2017carla}. Identifying and randomising relevant simulation parameters adds further to the complexity, making the entire process a resource-intensive endeavour. Additionally the randomisation process can occasionally lead to inaccurate and infeasible results.

To address the above mentioned issues, in this paper we introduce a novel approach to domain randomisation that leverages the rich visual diversity inherent in \textbf{video games}; see Fig. \ref{fig:alignmentdemo}. Specifically our framework distinguishes itself by not relying on game engine access for the randomisation process. This unlocks the potential of CV to be trained and tested on videos from commercial-grade video game titles, a capability that has, until now, remained elusive. Our \underline{Beh}aviour \underline{A}lignment of \underline{V}ideo Game \underline{E}ncodings (\textbf{BehAVE}) framework harnesses the unique characteristic of gameplay videos opposed to any other videos available. Gameplay video footage is generated by sequential player actions (\textit{i.e.,} controller inputs) that control on-screen animations characterised as \textit{player behaviour}. Using player actions, BehAVE is able to align video encodings of similar player behaviours across visually diverse games (\emph{i.e.}, different \textit{domains}). Crucially, it employs \textit{semantic action encoding}, a method that infuses semantic information about behaviour through textual descriptions of actions, which are encoded using pre-trained text encoders \cite{patil2023survey}. As a result, the text encodings of player actions guide the behavioural alignment of video encodings.

We train BehAVE's alignment module, implemented on top of foundation video encoders \cite{tong2022videomae}, across a diverse array of games from the first person shooter (FPS) game genre, namely our introduced \emph{SMG-25} dataset. Our findings suggest that BehAVE is able to uncover similar behavioural patterns---despite visual distinctions, such as variations in game style or aesthetics---across unseen games of the SMG-25 dataset (see Figure \ref{fig:alignmentdemo} for an illustrative example). The alignment training proves efficient even with small datasets comprising only a few games, and it demonstrates robustness across various tested video and text encoders.

To assess the \emph{transferability} of BehAVE we test it on a 
% behaviour classification task across various FPS games, while solely 
video understanding task across various FPS games, while solely being trained on the FPS game \textit{Counter Strike: GO} (Valve, 2012). Further, we test a more challenging scenario evaluating the transfer performance to the FPS genre from \textit{Minecraft} (Mojang, 2011) a first person game from a different genre (non-FPS). Our findings indicate higher transferability when learning to classify behaviour from our aligned representation space as compared to without, showcasing up to $22\%$ higher classification accuracies across the different behaviour categories tested. We view this as a potential avenue for annotating extensive datasets of online gameplay videos with behaviour labels, thereby serving as a stepping stone towards learning generalised representations of behaviour in videos. Our contributions can be summarised as follows: (1) We introduce the \textbf{BehAVE} framework for domain randomisation via commercial video games; (2) We propose \textbf{Semantic Action Encoding} for representing player actions as textual descriptions processed through a pretrained text encoder; (3) We introduce the \textbf{SMG-25} dataset of synchronised gameplay and actions.
\section{Background}
\label{sec:relatedwork}

\textbf{Video Understanding in CV.} Video understanding methods seek to interpret visual information embedded within temporal image-sequences. Recent strides in deep learning have led to attaining remarkable performance in diverse video understanding tasks, including but not limited to \emph{video classification} \cite{bertasius2021space}, \emph{video summarisation} \cite{apostolidis2021combining}, \emph{short and long-form video understanding} \cite{wu2021towards}, and \emph{object tracking} \cite{zhu2023cross}. Current endeavours focus on training strategies that are independent of any specific downstream task. The resulting \emph{video foundation models} \cite{wang2023videomaev2, wang2023masked} yield powerful video representations, readily applicable across a diverse range of tasks. We use such foundation models in our study courtesy of their out-of-the-box performance and employ them as is (\emph{i.e.} frozen) bounded by limited computational resources \cite{togelius2024choose}. This underscores the computational efficiency of our video understanding framework, ultimately enhancing its accessibility.

\textbf{Transferable CV and Domain Randomisation.} Despite their impressive out-of-the-box performance, foundation models showcase limited capacities on transferring knowledge from one domain to another \emph{visually distinct} domain, primarily due to the \enquote{domain gap} challenge \cite{trivedi2021contrastive, trivedi2023towards}. Tobin et al. \cite{tobin2017domain} introduced the technique of \textit{domain randomisation} to train transferable vision models by injecting variability during learning. This is achieved by randomising the rendering of a simulator that generates training data. Leiprecht \cite{leiprecht2020using} showcases the efficacy of domain randomisation in CARLA \cite{dosovitskiy2017carla}, a large-scale driving simulator. Mishra et al. \cite{mishra2022task2sim}, however, bring to light the numerous complexities associated with identifying and tweaking relevant parameters of such simulators. Furthermore, Kim et al., \cite{kim2022transferable} emphasise the \emph{limited} variability that can be attained from a single simulator, impacting the transfer capacity \cite{yue2019domain}. Hence, in this work, we adopt a simulator-free approach for visual domain randomisation.

\textbf{Video Games for CV.} Inspired by insights from \cite{risi2020increasing} and \cite{reed2022generalist} suggesting that procedurally generated sets of diverse games enhance generality in machine learning, we explore the use of existing video games in CV. Several recent studies investigate the use of commercial-standard games as an alternative to dedicated in-lab simulators or procedural game level generation approaches, in an attempt to circumvent limitations related to inaccessible game engines. Notably, \textit{Grand Theft Auto 5} (Rockstar, 2013) serves as a popular video game for collecting annotated data, achieved by intercepting rendering communication between graphics hardware and the screen buffer \cite{richter2016playing, krahenbuhl2018free, taesiri2022clip} or employing a game modification such as \enquote{infrared vision mod} \cite{gu2023infrared}. Alternatively, Pearce and Zhu \cite{pearce2022counter} gather internal game state information from \textit{CS:GO} by probing the machine's memory. In contrast to such prior works involving the \enquote{reverse-engineering} of game engines, our approach captures high-level game information such as player actions using raw inputs from the machine's I/O devices, thereby simplifying the collection of annotated gameplay and boosting the scalability of our method across numerous commercial video games.

\textbf{Multimodal Alignment of CV Models.} Given that BehAVE considers different modalities of input such as videos of gameplay and corresponding player actions, we draw inspiration from contemporary work in \textit{video action recognition} \cite{zhu2020comprehensive}. In their work with a paired video-text caption dataset, Song et al. \cite{song2016unsupervised} extract \emph{verbs} from captions and use them as \emph{action labels}. Our framework builds on similar principles utilising language models \cite{patil2023survey}, but instead, encodes player actions; BehAVE then uses these action encodings for alignment with another modality, namely gameplay videos. To achieve this, we rely on \textit{multimodal alignment frameworks} that operate with and align vision and language such as CLIP \cite{radford2021learning} and VideoCLIP \cite{xu2021videoclip,ko2022video}. Drawing upon insights from the aforementioned studies, we propose a novel method for performing visual domain randomisation with commercial games by aligning gameplay videos with semantically represented player actions.

%Nemanja; this algorithm needs to be placed within section 3 (i.e. next page) - not just before it. (DONE)

\begin{algorithm}[!hbt]
	\caption{Behaviour-Alignment Training with BehAVE} 
	\begin{algorithmic}[1]
        \STATE \textbf{Inputs:} Games Dataset $\mathbb{D}$, semantic action mapper $m$, pre-trained text encoder $h$, pre-trained video encoder $f$ and trainable alignment projector $p$.
        \FOR {(video $V,$ actions $A$) in $\mathbb{D}$}
            \STATE Compute video encoding $z^{\text{video}}=f(V)$
            \STATE Compute action encoding $z^{\text{caption}}=h(m(A))$
            \STATE Project to aligned encoding $z^{\text{align}}=p(z^{\text{video}})$
            \STATE Calculate loss $\mathcal{L}_{\text{cos}}=1-cosine(z^{\text{align}},z^{\text{caption}})$
            \STATE Update projector network parameters $p_{\theta}$
        \ENDFOR
        \STATE \textbf{Output:} Trained alignment projector $p$. 
	\end{algorithmic}
 \label{alg:behave}
\end{algorithm}

\section{The BehAVE Framework} 
\label{sec:methodology}
\begin{figure}[!tb]
    \centering
    \begin{subfigure}[b]{\textwidth}
        \centering
        \includegraphics[width=\textwidth]{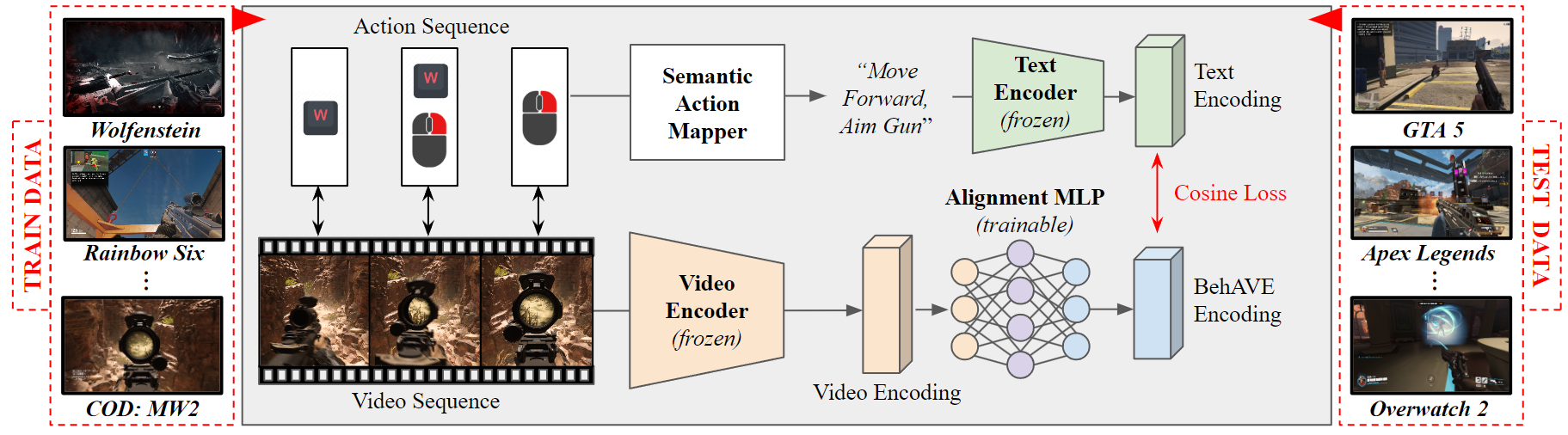}
        \caption{Behaviour Alignment}
        \label{fig:behavior_alignment}
    \end{subfigure}
    % \hfill
    \begin{subfigure}[b]{\textwidth}
        \centering
        \includegraphics[width=\textwidth]{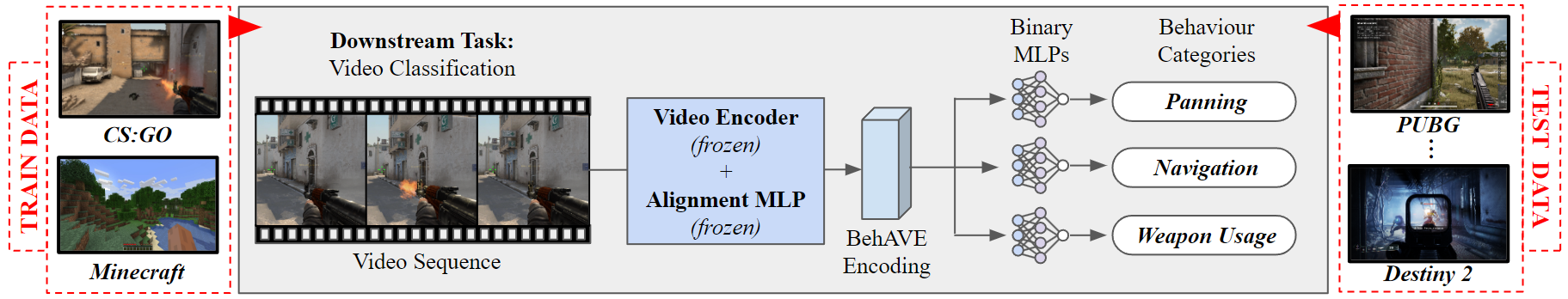}
        \caption{Behaviour Classification}
        \label{fig:behavior_classification}
    \end{subfigure}
    \caption{Overview of experiments and datasets used: \textbf{(a) Behaviour Alignment:} BehAVE is trained on synchronised gameplay video and player actions from the SMG-25 train dataset, and evaluated on unseen games from the SMG-25 test dataset. \textbf{(b) Behavior Classification:} We test the transferability of a \textit{video classification} task. BehAVE is trained independently on CS:GO and Minecraft, and transferred to the SMG-25 test dataset.}
    \label{fig:experiments}
\end{figure}

As introduced earlier, we present \emph{BehAVE}, a framework operating on paired visuals-and-actions datasets derived from commercial games, with the aim of aligning video encodings based on similar player behaviour. The BehAVE method is presented in Algorithm \ref{alg:behave} and visually depicted in Figure \ref{fig:behavior_alignment}. In Section \ref{subsec:drwithgames}, we explain the special structured dataset of games imperative for our framework, followed by Section \ref{subsec:videorepresentation} covering the encoder models used for both modalities. Finally, Section \ref{subsec:alignmenttraining} details the training method employed.

\subsection{Games Dataset for Training BehAVE}
\label{subsec:drwithgames}
A crucial component of our domain randomisation framework involves the meticulous preparation of a dataset adhering to a specific \emph{structure} that accommodates semantically similar visual content represented across diverse visual styles. We enforce this structure via the selection criteria of the various commercial games in the training dataset. Note that since BehAVE is trained upon player-game interaction data, we do not require access to the game engines, making it a viable strategy to use commercial games.

\textbf{Game Selection for Domain Randomisation.} Let $\mathcal{G}$ represent a game, with a frame-renderer $g$, and $\mathcal{G} \in \mathcal{C}$, where $\mathcal{C}$ denotes the family of games of a certain game genre category. Given that domain randomisation with customisable simulators involves the adjustment of simulator render parameters, we proceed to identify and formally define comparable parameters $\xi$ within the context of video games. Each game's renderer encompasses certain game-specific parameters denoted as \enquote{game style parameters} $(\xi^{\mathcal{G}})$, associated with either the visual aesthetics of the game, such as \textit{textures and colours of objects}, or the underlying rules governing the game, such as \textit{game physics}. These parameters are considered invariant throughout the game, reflecting game design choices made during development, and are less likely to be shared across all games of this genre category. In the context of our analysis involving multiple commercial games, we observe $\xi^{\mathcal{G}} \in \Xi$ where $\Xi$ represents the diverse \textit{global game-design space}, introducing implicit \enquote{randomisation} into our framework. This unique characteristic of games makes them ideal for the purpose of visual domain randomisation. 

Additionally, we also characterise all game-state-specific parameters of the renderer, including the \textit{player's spatial coordinates}, \textit{health or ammunition status}, and \textit{camera perspective}, as \enquote{game content parameters} $(\xi^{\mathcal{C}}_t)$, which dynamically evolve at each timestep $t$ in response to player interactions with the game environment. Note that these parameters remain largely consistent across different games that are categorised under the same genre.

\textbf{Synchronised Gameplay Recording.} At each timestep $t$, we obtain two synchronised information streams—visuals and actions. Player inputs or actions are selected from the shared action space of the game genre $\mathcal{C}$ and are represented by a set of $N^{\mathcal{C}}$ unique keypresses as $A_t=\{a_n\}_{n=1}^{N^{\mathcal{C}}}$, where $a_n \in \{0,1\}$. Visuals are recorded in the form of RGB frames $F_t \subset \mathbb R^{h\times w\times3}$ where $h$ is height and $w$ is width. The visuals of a game can be regarded as dynamic sequences of frames that arise from the interactions between the player and the game, as follows: $F_{t+1} = g(F_{t}, A_{t}, \xi_{t}^{\mathcal{C}} \mid \xi^{\mathcal{G}})$. Thus, the video frames are generated sequentially by the game renderer $g$ processing the game content and player action information at every timestep for the given predefined game style. This inherent characteristic of gameplay visuals, derived from player interactions, enables us to employ actions for effectively discerning visual content.

\textbf{Data Pre-processing.} Although we collect data at the timestep level, our framework operates on videos for identifying behaviour. To this end, we aggregate data over consecutive timesteps, forming a window of length $T$ to obtain video sequences $V=(F_1,F_2,...,F_{T})$ and action sequences $A=(A_1,A_2,...,A_{T})$. Consequently, for each game, the synchronised gameplay-actions dataset is denoted as $D^{\mathcal{G}} = \{(V_i,A_i)\}_{i=1}^{I}$ where $|D^{\mathcal{G}}| = I$. We construct the \emph{overall} dataset $\mathbb D = \bigcup_{\mathcal{G} \in \mathcal{C}} D^{\mathcal{G}}$ comprising of $k=|\mathbb D|$ \enquote{distinct} games from the game genre $\mathcal{C}$, adhering to the previously outlined selection strategy. This dataset forms the basis for behaviour alignment training.

% \begin{figure*}[t] 
%     \centering
%         \centering
%         \includegraphics[width=0.95\linewidth]{figures/alignment_recognition.png}
%     \caption{Outline of experiments and datasets used this study: \textbf{(a) Behavior Alignment:} BehAVE is trained on synchronized gameplay video and player actions from the SMG-25 train dataset, and evaluated on unseen games from the SMG-25 test dataset. \textbf{(b) Behavior Classification:} We test the transferability of a \textit{video classification} task. BehAVE is trained independently on CS:GO and Minecraft, and transferred to the SMG-25 test dataset.} 
%     \label{fig:experiments}
% \end{figure*}

\subsection{Encoding the Modalities}
\label{subsec:videorepresentation}
\textbf{Video Encoding and Alignment.} As previously stated, we harness the capabilities of a pre-trained video foundation model in our study for video understanding. Let $f$ denote the backbone model of a \textit{video encoder}; thus, the latent representation of the backbone's video encoding can be given by $z^{\text{video}}=f(V)$ (Algorithm \ref{alg:behave}, line 3), where $V$ denotes a 1 second video consisting of 16 consecutive frames. Note that we employ $f$ within our training in a frozen state, a decision influenced by our evaluation of foundational models as well as other computational constraints. To facilitate the alignment of representation spaces of different modalities within our framework, we employ a trainable MLP model $p$ that acts as an \textit{alignment projector} and operates on top of the video encoder, yielding the aligned projection encoding $z^{\text{align}}=p(z^{\text{video}})$ (Algorithm \ref{alg:behave}, line 5).

\textbf{Semantic Action Encoding.} As previously explained, each video is associated with a sequence of binary actions $A$, indicating the presence or absence of a key-press at every timestep. Binary labels for actions, however, offer limited insights into the inter-relationships among various sub-actions. For instance, in FPS games, the binary encodings of the four actions—\textit{move left}, \textit{move right}, \textit{shoot gun}, and \textit{aim gun}—are equidistant from one another. This encoding type, however, fails to capture the underlying semantic similarity between the first two actions (\textit{i.e.}, related to movement), the last two actions (\textit{i.e.}, related to weapon use) and also the semantic difference between these two behavioural categories.

To address the above limitation, we propose to equip BehAVE with a hand-crafted \textit{semantic action mapper} function $m$ which injects semantic information into the action encodings via text. It maps the binary sequence of keypresses to a behaviour text caption. Then, we use a pre-trained text foundation model in the form of a \textit{text encoder} $h$ that gives the caption's text encoding $z^{\text{caption}}=h(m(A))$ (Algorithm \ref{alg:behave}, line 4). We argue that such pre-trained encoders will be able to better capture the inter-relationships among the joint distribution of actions that are otherwise difficult to represent with binary action encodings.

\subsection{Alignment Training}
\label{subsec:alignmenttraining}
Upon obtaining the encodings of videos and actions, we initiate the training phase of the framework using the specified dataset of games. We observe different video sequences exhibiting similar behaviour across different games. Consequently, to align the representation space of the video encoder to match that of the text encoding of behaviour, we choose to train our projector head as attached to the video encoder. Subsequently, we use a loss $\mathcal{L}_{\text{cos}}$  (Algorithm \ref{alg:behave}, line 6) based on the cosine similarity between the video projector encoding and the behaviour text encoding so that the former aligns with the latter on the same (shared) representation space. The loss is defined as follows:

\begin{equation}
\mathcal{L}_{\text{cos}}(z^{\text{align}},z^{\text{caption}}) = 1 - \dfrac {z^{\text{align}} \cdot z^{\text{caption}}} {\left\| z^{\text{align}}\right\| _{2}\left\| z^{\text{caption}}\right\| _{2}}
\end{equation}

Upon completion of the alignment training, the video encoder equipped with the trained alignment projector can be utilised on any other visual content for video understanding, without requiring access to any other modalities---such as player actions---that only pertain to games. 

In summary, the introduced BehAVE framework operates as follows. The structurally enriched dataset of the framework facilitates domain randomisation, the pre-trained video encoder enables video understanding, the semantic action encoding introduces the semantic notion of behaviour, and the alignment training module ensures enhanced transferability in video understanding.

\section{Experiments}
\label{sec:expsetup}

Figure \ref{fig:experiments} outlines the two primary experiments conducted in our study: \ref{fig:behavior_alignment} \textit{Behaviour Alignment}, where we perform alignment training (Section \ref{subsec:behavioralignmentexp}), and \ref{fig:behavior_classification} \textit{Behaviour Classification}, where we assess transferability of the aligned models in a downstream classification task (Section \ref{subsec:behaviorclassificationexp}). Before delving into the experiments, in Section \ref{subsec:dataexp} we introduce the datasets and evaluation metrics employed. Note that all experiments have been carried out on a single GTX 1070 (8GB) GPU, highlighting the cost-effective nature of our method.

\subsection{Dataset and Metrics}
\label{subsec:dataexp}
\begin{figure}[!hbt]
    \centering
    \includegraphics[height=0.53\linewidth]{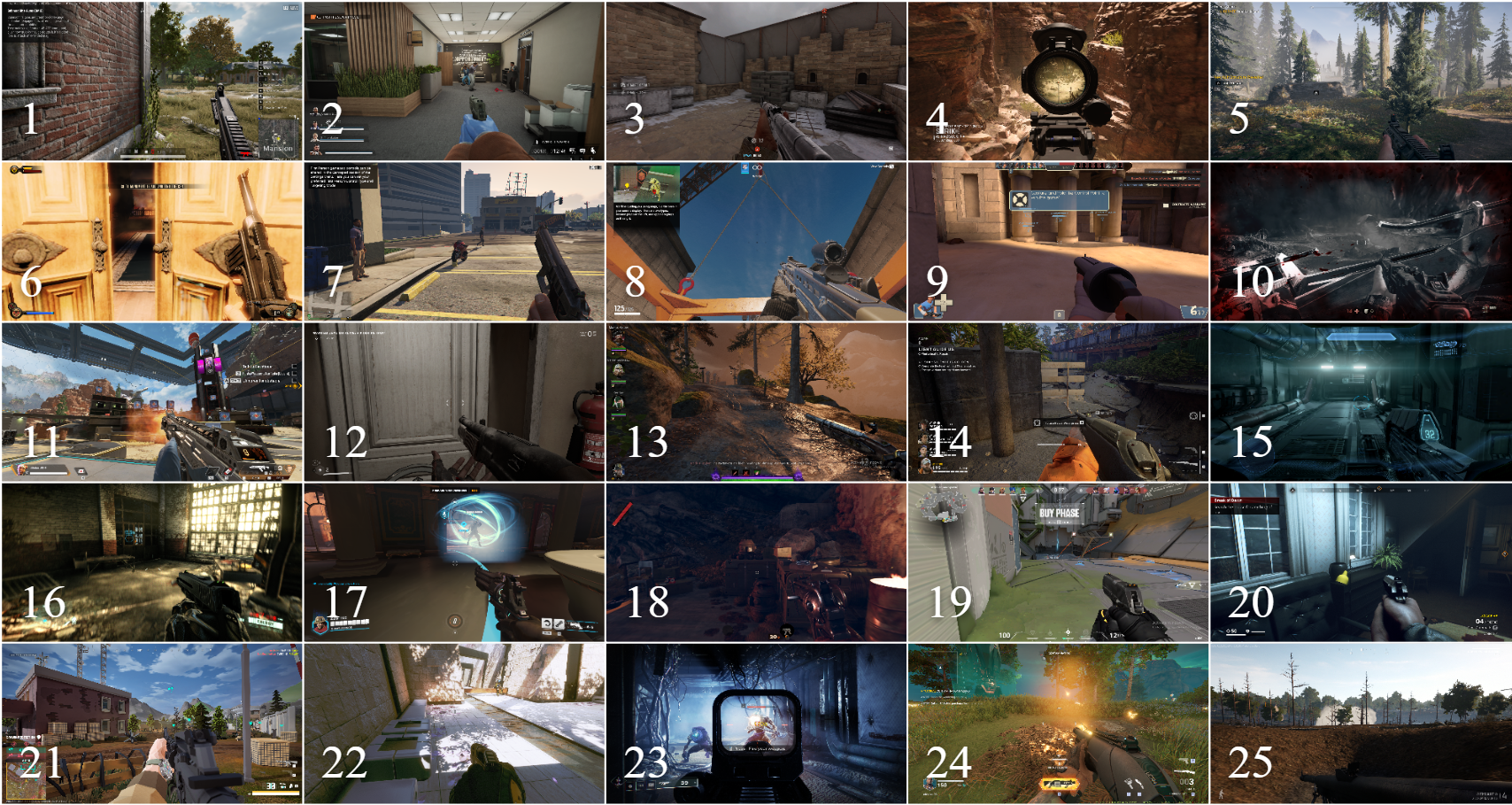}
    \caption{Screenshots from all games of the SMG-25 dataset: \textbf{1)} PUBG: Battlegrounds (\textit{PUBG Studios, 2017}); \textbf{2)} Payday 3 (\textit{Starbreeze Studios, 2023}); \textbf{3)} Insurgency: Sandstorm (\textit{New World Interactive, 2021}); \textbf{4)} Call of Duty: MW2 (\textit{Infinity Ward, 2022}); \textbf{5)} Far Cry 5 (\textit{Ubisoft, 2018}); \textbf{6)} Bioshock Infinite (\textit{Irrational Games, 2013}); \textbf{7)} Grand Theft Auto 5 (\textit{Rockstar, 2013}); \textbf{8)} Rainbow Six: Siege (\textit{Ubisoft, 2015}); \textbf{9)} Team Fortress 2 (\textit{Valve, 2007}); \textbf{10)} Wolfenstein (\textit{Machine Games, 2014}); \textbf{11)} Apex Legends (\textit{Respawn Entertainment, 2019}); \textbf{12)} Atomic Heart (\textit{Mundfish, 2023}); \textbf{13)} Warhammer: Vermintide 2 (\textit{Fatshark, 2018}); \textbf{14)} Back 4 Blood (\textit{Turtle Rock Studios, 2021}); \textbf{15)} Halo 4 (\textit{343 Industries, 2012}); \textbf{16)} Crysis 2 (\textit{Crytek, 2011}); \textbf{17)} Overwatch 2 (\textit{Blizzard Entertainment, 2022}); \textbf{18)} Deathloop (\textit{Arkane Lyon, 2021}); \textbf{19)} Valorant (\textit{Riot Games, 2020}); \textbf{20)} Generation Zero (\textit{Systemic Reaction, 2019}); \textbf{21)} Polygon (\textit{Readaster Studio, 2020}); \textbf{22)} Titanfall 2 (\textit{Respawn Entertainment, 2016}); \textbf{23)} Destiny 2 (\textit{Bungie, 2017}); \textbf{24)} Shatterline (\textit{Frag Lab, 2022}); \textbf{25)} Operation Harsh Doorstep (\textit{Drakeling Labs, 2023}).}
    \label{fig:fpsgames}
\end{figure}

We test BehAVE on three datasets, namely \textit{SMG-25}, \textit{CS:GO}, and \textit{Minecraft}, across the two experiments of our study. The \textbf{SMG-25} (\underline{S}ynchronised \underline{M}ulti-\underline{G}ame FPS Dataset) is our newly introduced dataset illustrated in Figure \ref{fig:fpsgames} that encompasses synchronised gameplay visuals and player action data from multiple commercial First Person Shooter (FPS) games, gathered following the structure outlined in Section \ref{subsec:drwithgames}. It comprises over $\sim$250K data points spanning 25 visually diverse FPS games, encompassing actions related to player behaviour categories such as \textit{panning} (player looking around), \textit{navigation} (player moving in the environment) and \textit{weapon usage} (player engaging the equipped weapon). We partition it into an SMG-25 \textit{train set} for use in Experiment (a) and an SMG-25 \textit{test set} used for evaluations in both Experiments (a) and (b). The train-test splits consist of disjoint sets of games, enabling evaluation of zero-shot performance on unseen games. Further details about this dataset are available in supplementary material. Additionally, we source similar gameplay-actions data from other commercial games, namely \textbf{CS:GO} (180K data points of \textit{cs-dust} level from \cite{pearce2022counter}) and \textbf{Minecraft} (50K data points of \textit{contractor demonstrations} from \cite{baker2022video}), which serve as training datasets for Experiment (b).

\textbf{Evaluation Metrics.} To comprehensively assess the alignment quality in Experiment (a) and the transferability of aligned models in Experiment (b), we employ several metrics. The \textit{Silhouette Score} \cite{rousseeuw1987silhouette}, ranging from -1 to 1, quantifies the cluster quality of embeddings. In Experiment (a), it is used to gauge the effectiveness of the alignment projection based on behaviour categories as cluster labels, with a higher score indicating better-defined clusters. Additionally, the \textit{Transferability Score} evaluates results in Experiment (b) by considering the percentage difference in classification test accuracy between models trained on BehAVE encodings and those trained on foundation video encodings. A positive difference signifies better transferability for the alignment method relative to the corresponding foundation method, and vice versa. 

\subsection{Behaviour Alignment}
\label{subsec:behavioralignmentexp}
In this experiment, depicted in Figure \ref{fig:behavior_classification}, we test the BehAVE framework on the SMG-25 dataset. Table \ref{tab:videoEncoders} provides details on various pre-trained video and text encoders analysed, while a comprehensive report on the configurations tested is available in supplementary material. For the alignment projector, we opt for a 4-layer MLP with ReLU activations and 50\% dropout. The size of the final layer of the MLP is adjusted to match the encoding size of the selected text encoder. As elaborated next, we perform a thorough analysis of various design choices incorporated into our framework.

\begin{table*}[!htb]
\caption{The video and text foundation models used in our experiments. ($\dag$) For video and text models, respectively, input size indicates timesteps of RGB frames and maximum token length. The number of parameters and the encoding size of the models are also listed.}
\label{tab:videoEncoders}
    \vskip 0.15in
    \centering
    \begin{small}
        \begin{sc}
        \resizebox{\textwidth}{!}{%
        \begin{tabular}{clccccl}
        \toprule
        \textbf{Input} & \textbf{PreTrain Method} & \textbf{Model} & \textbf{Input Size}\textsuperscript{$\dag$} & \textbf{\#Params} & \textbf{Encoding Size} \\ \midrule
         \multirow{3}{*}{Video} & I3D \cite{carreira2017quo} & 3D-ConvNet & $16\times3\times224\times224$ & 79M & 512 \\
         % \rule{0pt}{10pt}
         & VideoMAEv2 \cite{wang2023videomaev2} & \multirow{2}{*}{ViT-Base} & \multirow{2}{*}{$16\times3\times224\times224$} & \multirow{2}{*}{87M} & \multirow{2}{*}{768} \\
         & MVD \cite{wang2023masked} &  &  &  &  \\ \midrule
        \multirow{3}{*}{Text} & GPT-2 \cite{radford2019language} & \multirow{3}{*} {Transformer} & 512 tokens & 110M & 768 \\       
        & CLIP \cite{radford2021learning} &  & 77 tokens & 63M & 512 \\
        & BERT \cite{reimers-2019-sentence-bert} &  &  256 tokens & 33M & 384   \\
        \bottomrule
        \end{tabular}}
        \end{sc}
    \end{small}
    \vskip -0.1in
\end{table*}

\textbf{Action Encoding Schemes.} We conduct a comparative study on our semantic action encodings, focusing on actions alone, without videos, in our dataset. We compare the cluster quality of actions encoded traditionally in binary form (\textit{i.e.,} \textit{keypress labels}) to text encodings (\textit{i.e.,} \textit{behaviour captions}), with the aim of highlighting benefits of infusing semantics into our BehAVE framework. Results using only the unique set of actions from SMG-25 are presented in Section \ref{subsec:behavioralignment}.

\textbf{Impact of Alignment Training.} Subsequently, we train BehAVE with the previously mentioned pre-trained video and text encoders and analyse the benefits of aligning the representation space of the video encoder with that of the text encoder. The alignment projector is trained for 10 epochs using the \textit{adam} optimiser with a learning rate of $1e^{-3}$. The training data comprises a set of 15 games from SMG-25, processed in batch sizes of 128. The resulting representation space post-alignment is evaluated using the \textit{silhouette score} metric on 10 games from the SMG-25 test set, allowing us to assess the ``zero-shot'' performance of our methods on unseen games. In Section \ref{subsec:behavioralignment} we present results from 5 independent runs, with randomised train-test splits for each run. 

\textbf{Sensitivity Analysis of \textit{k}.} For practical applications it is important to analyse how the number of games (\textit{k}) used during training affects alignment performance. Thus, we explore the impact of varying the number of games included in the training set, ranging from 1 to 15 and compare them on a fixed test set of 10 games. We randomise the selection of games in the train and test sets across the 5 runs reported in Section \ref{subsec:behavioralignment}.

\begin{table*}[t]
  \begin{center}
  \caption{\textbf{Impact of Alignment Training.} Average silhouette scores (with standard deviations) over 5 runs for behaviour and game labels on the SMG-25 test set. Higher scores indicate better performance for behaviour categories, while lower scores indicate better performance for game labels. Best performing models are highlighted in bold.} %Nemanja: it is not clear what is bold in this figure - please clarify (DONE)
  \vskip 0.15in
  \centering
  \resizebox{\textwidth}{!}{%
  \begin{small}
    \begin{sc}
      \begin{tabular}{clc|cccc}
        \toprule
         \multicolumn{1}{c}{\textbf{Alignment}} & \multicolumn{1}{l}{\textbf{Encoder Methods}} & \multicolumn{1}{c|}{\textbf{Alignment}} & \multicolumn{3}{c}{\textbf{Behaviour Labels}} & \multicolumn{1}{c}{\textbf{Game}}  \\
         \multicolumn{1}{c}{\textbf{(Approach)}} & \multicolumn{1}{l}{\textbf{(Video - Action)}} & \multicolumn{1}{c|}{\textbf{Dimension}} & \textbf{Panning} $\uparrow$ & \textbf{Navigation} $\uparrow$ & \textbf{Weapon} $\uparrow$ & \multicolumn{1}{c}{\textbf{Label} $\downarrow$}  \\ \midrule
        \multirow{2}{*}{Foundation} & I3D - None & 512 & $0.03_{\pm0.04}$ & $0.04_{\pm0.04}$ & $0.00_{\pm0.01}$ & $0.07_{\pm0.07}$  \\
        \multirow{2}{*}{(Baseline)} & VideoMAEv2 - None  & 768 & $0.08_{\pm0.00}$ & $0.13_{\pm0.00}$ & $0.03_{\pm0.00}$ & $0.10_{\pm0.00}$\\
         & MVD - None & 768 & $0.14_{\pm0.02}$ & $0.09_{\pm0.04}$ & $0.01_{\pm0.01}$ & $-0.05_{\pm0.02}$  \\ 
        \midrule
        \multirow{2}{*}{Keypress} & I3D - Binary & 16 & $0.36_{\pm0.01}$ & $0.45_{\pm0.01}$ & $0.30_{\pm0.02}$ & $-0.21_{\pm0.01}$  \\
        \multirow{2}{*}{(Naive)} & VideoMAEv2 - Binary & 16 & $0.35_{\pm0.00}$ & $0.48_{\pm0.01}$ & $0.32_{\pm0.01}$ & $-0.16_{\pm0.00}$ \\
        & MVD - Binary & 16 & $0.43_{\pm0.02}$ & $0.44_{\pm0.03}$ & $0.09_{\pm0.01}$ & $-0.24_{\pm0.02}$ \\ \midrule
        \multirow{2}{*}{BehAVE}& VideoMAEv2 - GPT-2 & 768 & $\mathbf{0.51_{\pm0.02}}$ & $\mathbf{0.58_{\pm0.05}}$ & $0.20_{\pm0.05}$ & $\mathbf{-0.29_{\pm0.06}}$\\
        \multirow{2}{*}{(Ours)} & VideoMAEv2 - CLIP & 512 & $0.40_{\pm0.01}$ & $0.49_{\pm0.01}$ & $\mathbf{0.35_{\pm0.00}}$ & $-0.20_{\pm0.01}$ \\ 
        & VideoMAEv2 - BERT & 384 & $0.41_{\pm0.03}$ & $0.48_{\pm0.04}$ & $\mathbf{0.35_{\pm0.02}}$ & $-0.17_{\pm0.03}$  \\
        \bottomrule
      \end{tabular}
    \end{sc}
  \end{small}}
  \vskip -0.1in
  \label{tab:alignmentscores}
  \end{center}
\end{table*}

\subsection{Behaviour Classification}
\label{subsec:behaviorclassificationexp}
Following alignment training, we emphasise the benefits of BehAVE encodings for transferring a downstream task across visually distinct domains. For this purpose, we select video classification as a representative task in video understanding, with behaviour categories serving as the class labels. As depicted in Figure \ref{fig:experiments}b, for each behaviour category, we train a classifier (3-layer MLP with binary output) on video encodings of a single game (source domain) not included in our SMG-25 dataset, and evaluate the performance of this classifier on multiple FPS games from the SMG-25 test set (target domains). We first use foundation video encodings as input to this classifier, and then BehAVE encodings are employed in the same fashion. We ultimately compare the classifiers' performance on FPS games between the two methods using the \textit{transferability score} metric. We first perform this experiment with our source domain being CS:GO, a game of the \emph{same} genre, followed by a more challenging experiment with our source domain being Minecraft, a \emph{similar} first-person perspective game but from non-shooter genre. In both cases, we evaluate classification performance on the SMG-25 test set of games unseen in alignment training to gauge the zero-shot transfer capacity of our classifiers; results are reported in Section \ref{subsec:behaviorclassification}.
\section{Results}
\label{sec:results}

\subsection{Behaviour Alignment}
\label{subsec:behavioralignment}
\begin{figure}[!tb]
    \centering
    \begin{subfigure}[b]{0.5\textwidth}
        \centering
        \includegraphics[width=\textwidth]{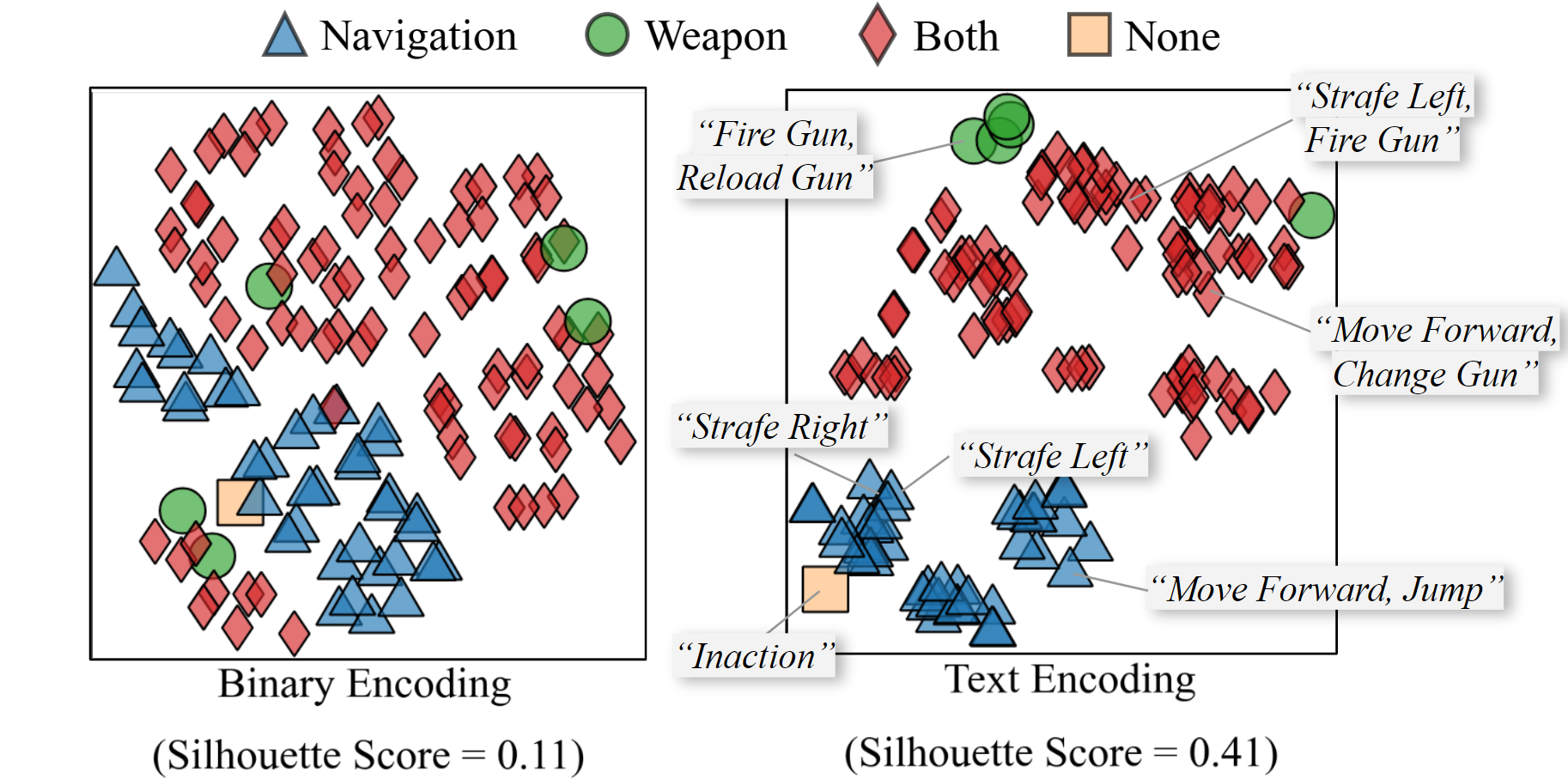}
        \caption{}
        \label{fig:textembeddings}
    \end{subfigure}
    % \hfill
    \begin{subfigure}[b]{0.49\textwidth}
        \centering
        \includegraphics[width=\textwidth]{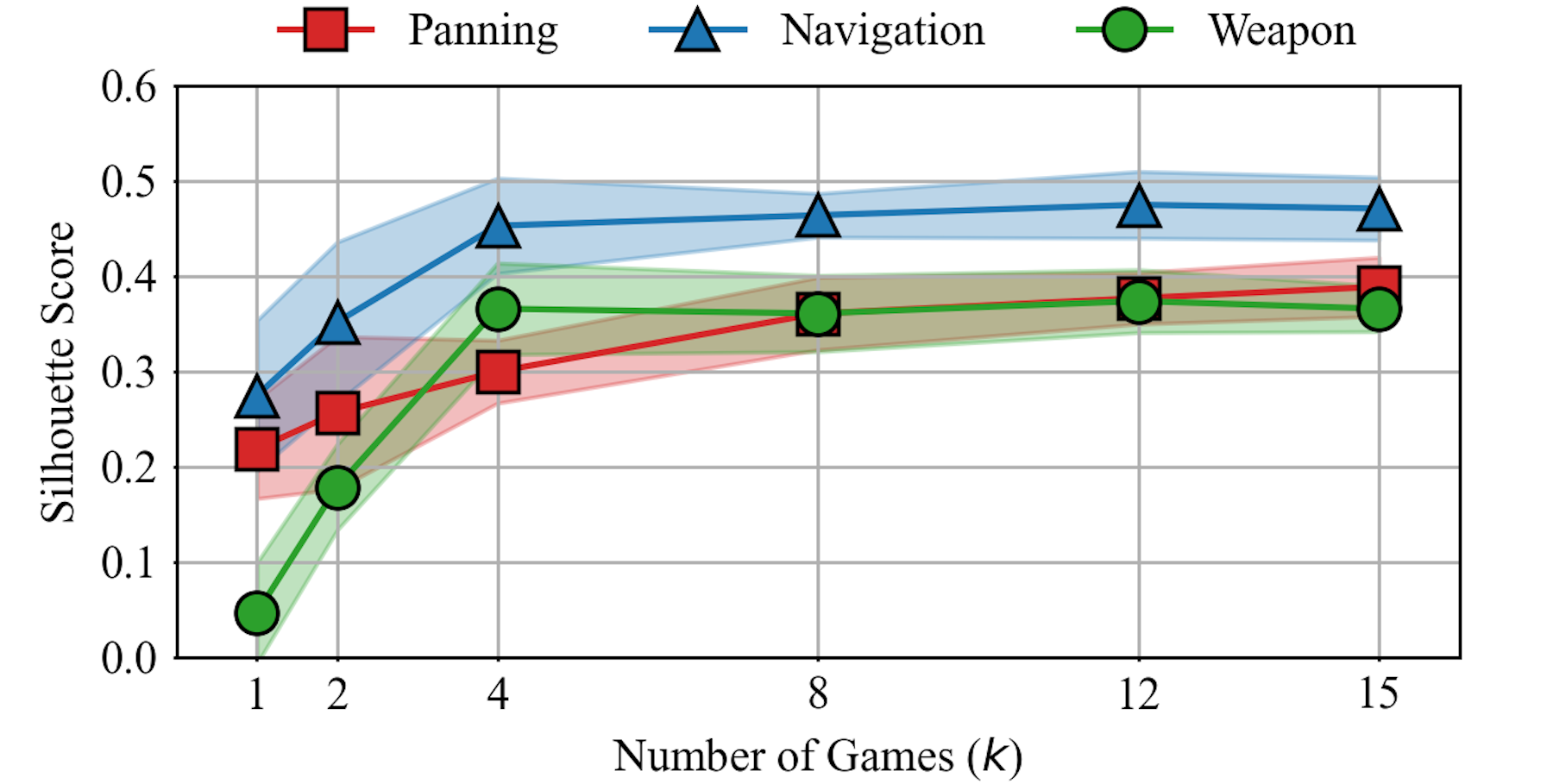}
        \caption{}
        \label{fig:noofgames}
    \end{subfigure}
    \caption{Behaviour alignment experiments: (a) t-SNE embeddings and corresponding silhouette scores of actions encoded as binary labels (left) compared to pretrained text encoders (right). (b) Effect of varying the number of games in alignment training on behaviour category clustering across 10 test games.}
\end{figure}

\textbf{Comparing Action Encoding Schemes.} Figure \ref{fig:textembeddings} presents the comparison between binary action encoding and semantic text encoding from the pre-trained CLIP text encoder \cite{radford2021learning}. The silhouette score for binary actions ($0.11$) is significantly lower than that for text encodings ($0.41$), showing that text encodings produce better cluster quality. Findings are further supported by t-SNE projections, where actions in the same behaviour categories appear to form more compact and distinct clusters. These results highlight the advantage of encoding actions as text based on semantics, rather than just as binary keypresses.
% The silhouette score for binary actions (0.11) is significantly lower than that of text encodings (0.41), indicating superior cluster quality for the latter. This advantage is also evident when encodings are projected onto two dimensions using t-SNE which shows that for actions belonging to the same behaviour categories, we observe more compact and well-separated clusters. This brings forth the benefit of encoding actions not merely as binary keypresses but rather as text encodings based on semantics.

\textbf{Impact of Alignment Training.} Table \ref{tab:alignmentscores} presents the analysis of the representation spaces before and after running BehAVE's alignment training. First, we observe that for all three behaviour categories, even alignment based on a naive approach of binary encoding of actions (see middle block of table) improves clustering quality across all video foundation models tested (upper block). Second, more interestingly, we observe even bigger improvements when using BehAVE's text encoding scheme (lower block). As a consequence of this alignment, we also observe that the domain gap, indicated by clustering on game labels, reduces for all alignment approaches compared to the foundation approach. The benefits of BehAVE are apparent across all tested configurations; a comprehensive list of all video and text encoders are provided in supplementary material.

\textbf{Sensitivity Analysis of \textit{k}.} Figure \ref{fig:noofgames} illustrates the sensitivity of the domain randomisation process to the number of games (\textit{k}) used in training. Performance varies across different behaviour categories. For \textit{weapon usage} and \textit{navigation}, we observe convergence between 4 to 6 games, indicating that a relatively small number of games is sufficient for identifying these categories. Surprisingly, for \textit{panning}, alignment continues to improve beyond 10 games. We argue that this phenomenon is attributed to the substantial variability in panning actions across different games, particularly due to different mouse sensitivity presets in SMG-25 that were approximated through visual inspection rather than precise extraction from the game engine (additional details in supplementary material). As a result, in our experiments with SMG-25, we chose 15 games for training and 10 for testing to balance the effectiveness of domain randomisation while maintaining a sufficiently large test set for zero-shot performance evaluation.

\subsection{Behaviour Classification}
\label{subsec:behaviorclassification}

\begin{figure}[!tb]
    \centering
    \begin{subfigure}[b]{\textwidth}
        \centering
        \includegraphics[width=\textwidth]{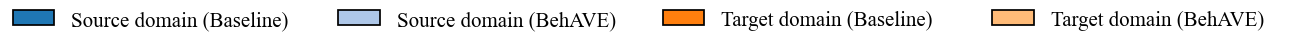}
    \end{subfigure}
    \begin{subfigure}[b]{0.49\textwidth}
        \centering
        \includegraphics[width=\textwidth]{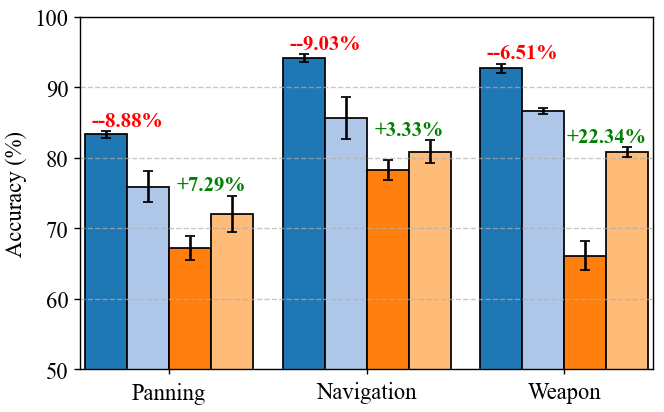}
        \caption{CS:GO (Source) Unseen games (Target)}
        \label{fig:res_cs_go}
    \end{subfigure}
    \begin{subfigure}[b]{0.49\textwidth}
        \centering
        \includegraphics[width=\textwidth]{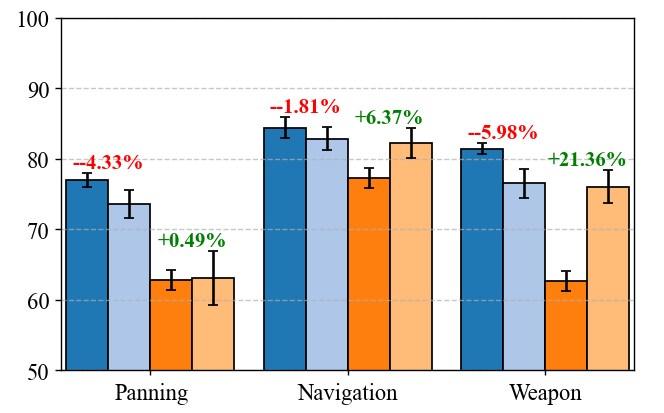}
        \caption{Minecraft (Source) Unseen games (Target)}
        \label{fig:res_minecraft}
    \end{subfigure}
    \caption{\textbf{Behaviour classification} accuracy across 3 behaviour categories when transferring from (a) CS:GO (FPS game) and (b) Minecraft (non-FPS game) to unseen FPS games. Although BehAVE (aligned) encodings perform slightly worse on source domain test sets than foundation (unaligned) encodings, they show significant improvements in generalisation to target domains, highlighting BehAVE's enhanced transfer capacities.}
    \label{fig:generalization}
\end{figure}

\textbf{Transfer from CS:GO (\textit{Same} Genre).} Figure \ref{fig:res_cs_go} shows transferability scores across all three behaviour categories, with CS:GO as the source domain and multiple unseen FPS games from the SMG-25 test set as target domains. Notably, we notice the poor transfer capacity of classifiers based on foundation encodings. In the case of BehAVE, while the absolute performance of the classifiers on source domains declines (highlighted in red above the respective bars), we observe a 3\% to 22\% improvement in transfer to target domains (highlighted in green). These findings are in line with contemporary studies, such as those in \cite{arjovsky2019invariant, wenzel2022assaying}, suggesting that minimising training error leads machines to absorb all correlations, including spurious ones from data biases. As shown in Figure \ref{fig:demo_foundation}, foundation models retain game-specific style information, leading to the exploitation of spurious correlations and poor generalisation. Conversely, BehAVE reduces data bias impact, enabling generalisation to new test distributions by discarding game-specific information, as seen in Figure \ref{fig:demo_behave}.

\textbf{Transfer from Minecraft (\textit{Similar} Genre).} Figure \ref{fig:res_minecraft} showcases similar benefits when using Minecraft as the source domain, despite its non-shooter nature. While \textit{panning} category obtains minimal improvement, we see a 6\% to 22\% improvement in classifying \textit{navigation} and \textit{weapon} categories. Although the weapon equipped by the Minecraft player is a pick-axe, the classifier is still able to transfer this knowledge to the gun-related behaviours of SMG-25 games.
%This finding is surprising since the weapon equipped by the Minecraft player is a pick-axe, but the classifier is still able to transfer this knowledge to the gun-related behaviours of SMG-25 games. 
This indicates that our model, trained on a diverse dataset covering multiple game genres with aligned semantic behaviour patterns, can generalise to new genres with overlapping characteristics. However, while the results demonstrate BehAVE's transferability across genres, this exploration is preliminary; further investigation into cross-genre transferability is left for future work.
%This indicates that our model, trained on a diverse dataset encompassing multiple game genres with aligned semantic behaviour patterns, can generalise its knowledge to new game genres with overlapping behavioural characteristics. It is crucial to acknowledge, however, that while the obtained results already showcase the transferability capacity of BehAVE even across genres, our exploration in this direction is preliminary; further investigation on across-genre transferability is deferred to future work.
\section{Discussion}
\label{sec:discussion}
% \textbf{Downstream Applications.} BehAVE's zero-shot performance presents promising applications in imitation learning, demonstrated across both first-person shooter and non-shooter game genres. Future experiments could explore training on multiple genres and measuring transferability to entirely different, unseen genres. While our focus has been on transferable video understanding, a complementary study could delve into representation learning optimized for specific downstream tasks. Fine-tuning foundation models might balance performance and transferability, tailoring BehAVE for specific applications. Supplementary material showcases some intial experiments delving into BehAVE's potential in learning inverse dynamics models.

\textbf{Downstream Applications.} BehAVE's zero-shot performance shows promising applications in imitation learning across both first-person shooter and non-shooter game genres. While our focus has been on transferable video understanding, a complementary study could explore representation learning optimised for specific downstream tasks. Supplementary material includes initial experiments on BehAVE's potential in learning inverse dynamics models. %BehAVE's zero-shot performance presents promising applications in imitation learning, demonstrated across both first-person shooter and non-shooter game genres. While our focus has been on transferable video understanding, a complementary study could delve into representation learning optimised for specific downstream tasks. Supplementary material showcases some initial experiments delving into BehAVE's potential in learning inverse dynamics models.

\textbf{Future Work.} Future experiments could employ BehAVE to various game genres, such as fighting, platformer, and driving games, with some, like driving games, requiring additional action preprocessing like discretising steering \cite{xu2017end}. Experiments could also assess training across multiple genres and evaluate transferability to unseen ones. Another direction is to keep pace with advancements in foundation models like large language and vision-language Models, deepening action understanding. With more computational resources and a larger team \cite{togelius2024choose}, fine-tuning these models could enhance performance and transferability, making BehAVE more adaptable to specific applications. %Future experiments could explore applying BehAVE to various game genres, such as fighting, platformer, and driving games. Some genres, like driving games, might require additional action preprocessing, such as discretising steering \cite{xu2017end}. Experiments could also assess training on multiple genres and evaluating transferability to entirely different, unseen ones. Another promising direction is to keep pace with advancements in foundation models, such as Large Language Models (LLMs) and Vision-Language Models (VLMs), which could deepen the understanding of analyzed actions. With more computational resources and a larger team of annotators, fine-tuning these models could balance performance and transferability, making BehAVE more adaptable to specific applications.

\textbf{Scalability.} As outlined in Section \ref{sec:expsetup}, we operate within a limited compute and with a small team of annotators, hindering data collection expansion and comparisons with end-to-end trained models. BehAVE could potentially perform better with increased data and computational resources; however, our focus is on introducing an accessible method that proves effective even at a smaller scale \cite{togelius2024choose}.

\textbf{Ethical impact.} This paper introduces a dataset of fully anonymised annotated FPS gameplay videos. Both the gameplay videos and the annotations were collected in-house from participants in a laboratory setting under a data collection protocol approved by the University Research Ethics Committee of the University of Malta. The protocol was inspired by previously peer-reviewed work cited in Section \ref{sec:relatedwork}. We acknowledge the emerging trends in machine learning for autonomous weapons research \cite{simmons2024ai} and wish to clarify that this project has no military applications and was not funded by military sources. Additionally, the data collection protocol is exclusively applicable to video games, and the dataset aims to support research in video games and artificial intelligence.

\textbf{Dataset} The dataset utilised in this research was derived from various commercial games. All game data consisting of screen captured game graphics, is the intellectual property of their respective game developers and publishers (see Figure \ref{fig:fpsgames}). To the best of our knowledge the inclusion of the data in this study falls under the provisions of fair use for the purpose of academic research, analysis, and non-commercial study under US law, and text and data mining (TDM) for research purposes under EU law. The dataset will be publicly available for future studies under a non-commercial license to ensure scientific reproducibility. %The dataset will be made publicly available for future studies to ensure scientific reproducibility under a non-commercial license.

\section{Conclusion}
\label{sec:conclusion}
% In this paper, we introduced BehAVE, a video understanding framework that employs a simulator-free domain randomisation technique, exploiting the inherent variations in graphical and animation styling found in commercial video games. Tested extensively within games of the first-person shooter genre, BehAVE is able to successfully align video encodings (displaying similar player behaviour) across visually distinct games. Compared to foundation models, BehAVE demonstrates significantly higher levels of zero-shot transferability to unseen FPS games in a downstream behaviour classification task when trained independently on a game from the same genre (CS:GO) and even a game from a different genre (Minecraft). The introduced BehAVE framework showcases superior zero-shot transferability capacities offering a robust and efficient method towards generalising perception across visually diverse environments.
In this paper, we introduced BehAVE, a video understanding framework that uses a simulator-free domain randomization method, leveraging the inherent variations in graphics and animations in commercial video games. Tested on first-person shooter games, BehAVE effectively aligns video encodings of similar player behaviours across different games. It outperforms foundation models in zero-shot transferability to unseen FPS games during a behaviour classification task, even when trained on games from different genres like Minecraft. BehAVE offers a strong and efficient method for generalising perception across visually diverse environments.

\section*{Acknowledgements}
We would like to express our gratitude to Tim Pearce for insightful discussions that contributed to the development of this work. Special thanks also go to Roberto Gallotta and Marvin Zammit for their invaluable assistance in collecting the necessary data for our research. Makantasis was supported by Project ERICA (GA: REP-2023-36) financed by the Malta Council for Science \& Technology (MCST), for and on behalf of the Foundation for Science and Technology, through the FUSION: R\&I Research Excellence Programme. Rasajski, Liapis and Yannakakis were supported by Project OPtiMaL funded by MCST through the SINO-MALTA Fund 2022.

\bibliographystyle{splncs04}
\bibliography{main}
\end{document}

% \title{Supplementary Material for \\ BehAVE: Behaviour Alignment of Video Game Encodings} 

\title{Supplementary Material for BehAVE} 



\author{Nemanja Rašajski\inst{1}\thanks{Equal contribution}\orcidlink{0009-0000-3487-1339} \and
Chintan Trivedi\inst{1}*\orcidlink{0000-0003-2749-2618} \and
Konstantinos Makantasis\inst{2}\orcidlink{0000-0002-0889-2766} \and 
Antonios Liapis\inst{1}\orcidlink{0000-0001-5554-1961}
\and
Georgios N. Yannakakis\inst{1}\orcidlink{0000-0001-7793-1450}}

\authorrunning{Rašajski et al.}

\institute{Institute of Digital Games, University of Malta, Malta 
\email{\{nemanja.rasajski,ctriv01,antonios.liapis,georgios.yannakakis\}um.edu.mt}\\
\and
AI Department, University of Malta, Malta\\
\email{konstantinos.makantasis@um.edu.mt}}

\maketitle

\section{The \emph{SMG-25} Dataset}
\label{appsec:smg25}

In this section we provide additional information detailing the entire pipeline of collecting, pre-processing and curating the SMG-25 dataset introduced in Section 3. Note that we intend to make this dataset publicly available upon the acceptance of the paper. 

\subsection{Data Collection}
\label{appsubsec:smg25datacollection}
\textbf{Hardware and software specifications:} Our data collection was carried out on a handheld gaming machine running Windows 11 operating system, equipped with an AMD 780M iGPU with 4GB of VRAM. 25 different commercial games were sourced from various game distribution platforms such as Steam (\emph{Valve, 2003}), Xbox Game Pass (\emph{Microsoft, 2017}) and Ubisoft Connect (\emph{Massive, 2020}). We run all games in full-screen mode on an external FullHD monitor connected to the handheld machine, along with keyboard and mouse attachments for player inputs. While running the game on the monitor screen, we visualise the collected data in real-time on the handheld's screen to ensure good quality and correctness of the data being recorded.

\textbf{Annotators:} Annotations were collected in-house by students and researchers of University of Malta in a laboratory setting using the previously mentioned hardware. Five annotators with varying skill levels participated in this study. Since player skill levels were not the focus of this study, no specific experience beyond basic gameplay ability was required.

\textbf{Game selection criteria:} We ensure that the games selected for inclusion in the dataset are listed under the first person shooter category on their respective distribution platforms, and that we do not select games whose content policy might be violated by collecting gameplay information. Games that run anti-cheat software in the background block the reading of mouse or keyboard information and were subsequently not included. Special attention was paid to ensure coverage of diverse yet representative games of the FPS genre, not only from a graphical styling or level of photo-realism standpoint, but also from the gameplay mode perspective (refer \enquote{Game Mode} column in Table \ref{tab:games_details}). Notably, within the FPS genre, there are certain gameplay modes which yield significant differences in the distribution of actions between games, influencing different player behaviours. For example, games played in the \enquote{Battle Royale} game mode usually involve significantly more navigation and environment exploration, and relatively less combat compared to games from the \enquote{Death Match} game mode. These differences are highlighted in the \enquote{Data Frequency by Behaviour} column in Table \ref{tab:games_details}-- for example in case of PUBG played in \enquote{Battle Royale} mode, only $24\%$ of the recorded data involves video sequences where a weapon was engaged, in contrast to $55\%$ in case of Back 4 Blood played in \enquote{Survival} mode.

\begin{table}[!tb]
  \centering
  \caption{Additional metadata and statistics of all games present in the study, including (1) 25 games from the newly introduced SMG-25 dataset, (2) Counter Strike: Global Offensive (\textit{Valve, 2012}) data sourced from \cite{pearce2022counter} and (3) Minecraft (\textit{Mojang, 2011}) data sourced from \cite{baker2022video}. Note that the impact of various game modes on data frequency statistics is detailed in Section \ref{appsubsec:smg25datacollection}, whereas the mouse-related metadata is detailed in Section \ref{appsubsec:dataprocessing}.}
    \vskip 0.15in
        \begin{sc}
        \resizebox{\textwidth}{!}{%
\begin{tabular}{lcc|ccc|cccc} \toprule
 \multicolumn{3}{c|}{\textbf{Data Collection}} & \multicolumn{3}{c|}{\textbf{Data Frequency (\%) by Behaviour}} & \multicolumn{4}{c}{\textbf{Mouse Recording}} \\
 \multicolumn{1}{c}{\textbf{Game Name}} & \multicolumn{1}{c}{\textbf{Game Mode}} & \multicolumn{1}{c|}{\textbf{\#Frames}} & \multicolumn{1}{c}{\textbf{Panning}} & \multicolumn{1}{c}{\textbf{Navigation}} & \multicolumn{1}{c|}{\textbf{Weapon}} & \multicolumn{1}{c}{\textbf{Type}} & \textbf{$\overline{\delta_x}$} & \textbf{$\overline{\delta_y}$} & \multicolumn{1}{c}{\textbf{$t(\delta)$}} \\ \midrule
PUBG & Battle Royale & 9768  & 82 & 96 & 24 & Free-form & 119 & 28 & 20 \\
Payday 3 & Bank Heist & 9056  & 96 & 90 & 43 & Free-form & 127 & 32 & 10 \\
Insurgency & Death Match & 10055  & 91 & 92 & 25 & Free-form & 75 & 13 & 20 \\
Call of Duty & Campaign & 10309  & 73 & 84 & 35 & Auto-center & 13 & 4 & 2 \\
Far Cry 5 & Free Roam & 10039  & 93 & 80 & 43 & Auto-center & 22 & 4 & 2 \\
Bioshock & Campaign & 10150  & 90 & 94 & 35 & Free-form & 174 & 40 & 20 \\
GTA 5 & Free Roam & 10046  & 83 & 95 & 19 & Free-form & 67 & 13 & 20 \\
Rainbow Six & Tutorial & 10153  & 64 & 78 & 21 & Free-form & 34 & 12 & 20 \\
Team Fortress 2 & Death Match & 11388  & 71 & 90 & 41 & Auto-center & 7 & 5 & 2 \\
Wolfenstein & Campaign & 10400  & 89 & 85 & 34 & Free-form & 75 & 18 & 20 \\
Apex Legends & Battle Royale & 11005  & 82 & 95 & 16 & Auto-center & 9 & 3 & 1 \\
Atomic Heart & Campaign & 10063  & 91 & 90 & 9 & Free-form & 72 & 18 & 20 \\
Warhammer & Survival & 11030  & 93 & 98 & 42 & Auto-center & 71 & 9 & 2 \\
Back 4 Blood & Survival & 10054  & 95 & 84 & 55 & Free-form & 102 & 31 & 20 \\
Halo 4 & Campaign & 10383   & 91 & 87 & 36 & Free-form & 125 & 29 & 20 \\
Crysis 2 & Campaign & 10430  & 84 & 90 & 32 & Free-form & 58 & 13 & 20 \\
Overwatch 2 & Death Match & 10888 & 89 & 93 & 45 & Auto-center & 43 & 8 & 2  \\
Deathloop & Campaign & 10669  & 96 & 90 & 21 & Free-form & 228 & 35 & 20  \\
Valorant & Death Match & 11011 & 89 & 90 & 26 & Free-form & 131 & 39 & 20 \\
Generation Zero & Campaign & 10511  & 86 & 91 & 22 & Free-form & 54 & 15 & 20 \\
Polygon & Death Match & 10288  & 81 & 92 & 23 & Free-form & 49 & 9 & 20\\
Titanfall 2 & Tutorial & 10542  & 80 & 87 & 34 & Auto-center & 8 & 3 & 2 \\
Destiny 2 & Campaign & 10467  & 93 & 90 & 32 & Free-form & 63 & 14 & 10 \\
Shatterline & Survival & 10250  & 95 & 91 & 38 & Free-form & 128 & 27 & 20 \\
Harsh Doorstep & Death Match & 10831  & 76 & 75 & 48 & Free-form & 48 & 9 & 20\\ \midrule
CS:GO & Death Match & 180000  & 88 & 83 & 36 & Auto-center & 72 & 18 & 20 \\\midrule
Minecraft & Free Roam & 50445  & 66 & 69 & 50 & Free-form & 98 & 19 & 40 \\ 
  \bottomrule
\end{tabular}}
        \end{sc}
    
    \vskip -0.1in
  \label{tab:games_details}
\end{table}

\textbf{Collection script:} We execute a python script in parallel to playing these games by self (\emph{i.e.}, by a human player and not a bot). In order for our script to interface with these close-sourced commercial games, we bypass the need for access to their respective game engines by instead, interfacing directly with the machine's input devices (mouse and keyboard; no gamepads were used) using python's \textit{Win32API} package as well as the machine's output device (screen) using \textit{OpenCV}. Since our study primarily focuses on analyzing gameplay footage, we implement real-time pause/resume feature in our script to ensure that only gameplay sequences are recorded and non-gameplay sequences such as \textit{loading screens}, \textit{cut-scenes} or \textit{menu overlays} are omitted to the best of our ability.

This script records screen data (RGB frames) and player actions (\emph{i.e.}, mouse movements, mouse clicks and keyboard keypresses) synchronized by timestamps, sampled at roughly 16Hz. Inspired by \cite{pearce2022counter}, our aim was to set the sampling frequency exactly at 16Hz; the actual capture frequency, however, was contingent upon hardware capability at our disposal, adjusting to accommodate the computational resources required for running graphically demanding games. As a result, data of some compute-intensive games were sampled approximately in the range between 12Hz and 16Hz instead. Screen captures of the game are taken at a resolution of $1920\times1080$, subsequently downsized to $398\times224$ before saving to disk due to storage limitations. Simultaneously, keyboard and mouse clicks are recorded as binary data (\emph{i.e.}, 1 for key-pressed state and 0 otherwise) and mouse movements are recorded in the form of positional $x$ and $y$ screen coordinates.

The SMG-25 dataset collected using this script contains $\sim$250K samples of synchronised frames and actions across the 25 FPS games, amounting to $\sim$6hrs of gameplay footage. For each game included in the dataset Table \ref{tab:games_details} provides a number of statistics of the data collected. The detailed list of all actions recorded and the frequency with which they appear in our dataset can be found in Table \ref{tab:recordeddata}. The specific behavioural actions and their corresponding behavioural category are selected in this study as they act as primitive and general action patterns prevalent in most, if not all, games of the FPS genre. Note that while we use the aforementioned script to collect data for FPS games in this paper, the script can potentially be employed without any modification to collect similar gameplay information for any other game genre. Future enhancements to our data collection pipeline may include sampling of data at higher frequency and integrating additional modalities such as sound.
 
\begin{table}[!tb]
    \centering
    \caption{Detailed list with description and statistics of all actions recorded by our script, as well as various animation-related hyper-parameters chosen for pre-processing these actions (detailed explanations in Sections \ref{appsubsec:dataprocessing} and \ref{appsubsec:labelstotext}).}
    \vskip 0.15in
        \begin{sc}
                \resizebox{\textwidth}{!}{%
            \begin{tabular}{cc|lc|c|ccc}
                \toprule
                  \multicolumn{2}{c|}{\textbf{Script Recording}} & \multicolumn{2}{c|}{\textbf{FPS Genre Behaviour}} & \multicolumn{1}{c|}{\textbf{SMG-25}} & \multicolumn{3}{c}{\textbf{Animation (\#frames)}}  \\
                \textbf{Device} & \textbf{Action/Key} & \multicolumn{1}{c}{\textbf{Description}} & \textbf{Category} & \textbf{Freq. (\%)} & \textbf{Delay} & \textbf{Length} & \textbf{Cutoff}  \\
                \midrule
                 \multirow{6}{*}{Mouse} & Move Left  & Look Left & Panning & 62 & 1 & 2 & 2 \\
                 & Move Right  & Look Right & Panning & 62 & 1 & 2 & 2 \\
                & Move Up  & Look Up & Panning & 36 & 1 & 2 & 2\\
                 & Move Down  & Look Down & Panning & 40 & 1 & 2 & 2\\
                 & Left click  & Shoot & Weapon & 19 & 1 & 2 & 2 \\
                 & Right click  & Aim & Weapon & 15 & 1 & 2 & 2 \\ 
                 \midrule
                \multirow{13}{*}{Keyboard} & W  & Move Forward & Navigation & 77 & 1 & 2 & 2\\
                & A  & Strafe Left & Navigation & 39 & 1 & 2 & 2\\
                & S  & Move Backward & Navigation & 11 & 1 & 2 & 2\\
                & D  & Strafe Right & Navigation & 40 & 1 & 2 & 2\\
                & R  & Reload & Weapon & 7 & 3 & 16 & 6\\
                & Space & Jump & Navigation & 8 & 1 & 2 & 2\\
                & L. Shift (Hold) & Sprint & Navigation & 20 & 1 & 2 & 6\\
                & L. Ctrl (Toggle) & Crouch & Navigation & 2 & 1 & 2 & 2\\
                & C (Toggle) & Crouch & Navigation & 0.1 & 1 & 2 & 2\\
                & 1 & Switch Item & Weapon & 0.4 & 3 & 8 & 6\\
                & 2 & Switch Item & Weapon & 0.4 & 3 & 8 & 6\\
                & 3 & Switch Item & Weapon & 0.1 & 3 & 8 & 6\\
                & F  & Interact & - & 8 & 3 & 5 & 3\\
                \bottomrule
            \end{tabular}}
        \end{sc}
    \vskip -0.1in
    \label{tab:recordeddata}
\end{table}

\subsection{Data Pre-Processing}
\label{appsubsec:dataprocessing}
In this section we highlight the necessary steps taken to convert raw action keypress values (obtained from our script) to animation labels. Note that these pre-processing steps are pertaining specifically to the FPS genre of games and may not be directly applicable to data collected for other game genres. 

\begin{figure}[!tb]
  \centering
  \begin{subfigure}[]{\columnwidth}
    \includegraphics[clip,width=\columnwidth]{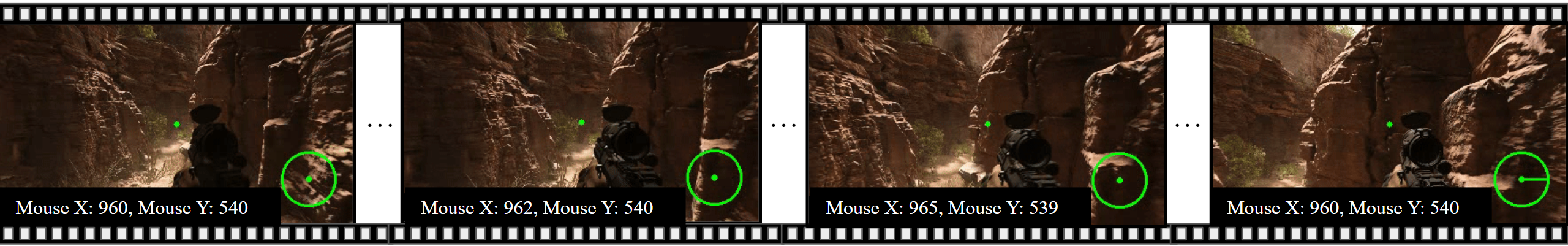}
    \caption{\textbf{Auto-Centre} type of mouse movement in \textit{Call of Duty: MW2}, highlighting the player-initiated mouse movement in the first three depicted frames and the system-initiated mouse coordinate reset in the final depicted frame.}
  \end{subfigure}
  
  \vspace{10pt}
  \begin{subfigure}[]{\columnwidth}
    \centering
    \includegraphics[clip,width=\columnwidth]{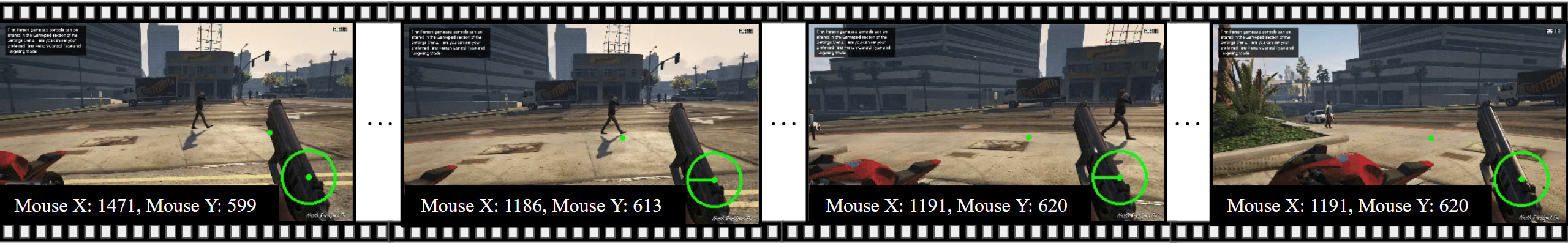}
    \caption{\textbf{Free-Form} type of mouse movement in \textit{Grand Theft Auto 5}, where all mouse movements are player-initiated and the mouse pointer can take any value within the boundaries of the screen.}
    \label{fig:mousemovements}
  \end{subfigure}
  \caption{Examples of the two types of mouse movements observed predominantly in FPS games. \enquote{Mouse X} and \enquote{Mouse Y} values displayed at the bottom left of each frame are the raw observations captured by the data collection script, also visualised on the frames by a green dot. The circular radar at the bottom right of each frame indicates the inferred direction of mouse movement. Notice the delay between the player moving the mouse and the associated panning animation occurring on the screen. For instance, in the case of \textit{Call of Duty: MW2}, the Mouse X value exhibits an initial increase over the first three depicted frames, yet the predominant portion of the \emph{pan-right} animation takes effect not until the final depicted frame.}
\end{figure}

\textbf{Mouse movements:} Mouse movements in FPS games are tied to the panning behaviour used by a player to look around the game environment. In this step we aim to convert our recorded mouse coordinates to mouse movement labels. Upon examining the raw mouse coordinate values, we identify two predominant methods employed by various FPS games in handling mouse movements— (1) \textit{Auto-Centre}: This involves the automatic reset of the pointer position to the centre of the screen once the user ceases mouse movement. (2) \textit{Free-Form}: In this mode, the mouse pointer can freely occupy any position within the bounds of the screen resolution. The reader is referred to Figure \ref{fig:mousemovements} for a visual representation of each scenario. The \enquote{Mouse Recording Type} column of Table \ref{tab:games_details} identifies the relevant category for each game. For Auto-Centre, we introduce a simple check that only player-initiated mouse movements are processed while the game-initiated pointer resets are ignored. No such checks are necessary for the Free-Form since all movements are player-initiated. Next, we use these raw coordinate values to infer a direction for mouse movement, categorised into one of the 8 possible discretised directions: \textit{up, down, left, right, up-left, up-right, down-left} and \textit{down-right}. For this purpose, we perform the following pre-processing operation.

\textbf{Delta Thresholding:} In order to correctly interpret mouse movement, we are required to consider that different games use different mouse sensitivity presets, resulting in significantly different deltas between consecutive timesteps (\emph{i.e.}, difference between successive mouse coordinate values) for different games. For each game, the average values of these deltas in horizontal direction ($\overline{\delta_x}$) and vertical direction ($\overline{\delta_y}$) are presented in the \enquote{Mouse Recording} column of Table \ref{tab:games_details}. Because of the nature of FPS games, players of this game genre tend to yield more drastic horizontal mouse movements compared to vertical movements \cite{pearce2022counter}. Furthermore, based on these deltas as well as fine-tuned adjustments based on human estimation of mouse sensitivity, we choose a minimum-threshold hyper-parameter ($t(\delta)$) to define significant panning animation visible on screen. These threshold values chosen for each game are provided in Table \ref{tab:games_details} and are applied to both horizontal and vertical axes of mouse movements for defining the overall panning direction.

\textbf{Animation Labels from Raw Actions:} A significant challenge we are faced with when we attempt to associate every single frame of the game with an appropriate animation label stems from the fact that player behaviour is not available directly from the game engine per se but rather it has to be inferred from raw recordings of the machine's input devices. Every keypress action taken by a player using an input device is sent to the game engine, which subsequently decides what animation to trigger on-screen, if any at all. One might claim that some of our labels are approximate \enquote{proxy-label} of on-screen animations in the form of keypresses. We are unable to directly address the issue of refining these \enquote{proxy-labels} as that would require either access to the game engines for accurate labelling, or significant human-resources for the manual labelling of all frames in our dataset. We consider both of these options as logistically infeasible and prohibiting in terms of cost and effort. We instead attempt to address such challenges through the framework introduced in this paper. We wish our method to be zero-shot transferable and accessible for domain randomisation and we design it with such limitations in mind: no access to the games themselves and no resources for manual action labelling. 

\begin{figure}[!tb]
  \centering
   \includegraphics[clip,width=\columnwidth]{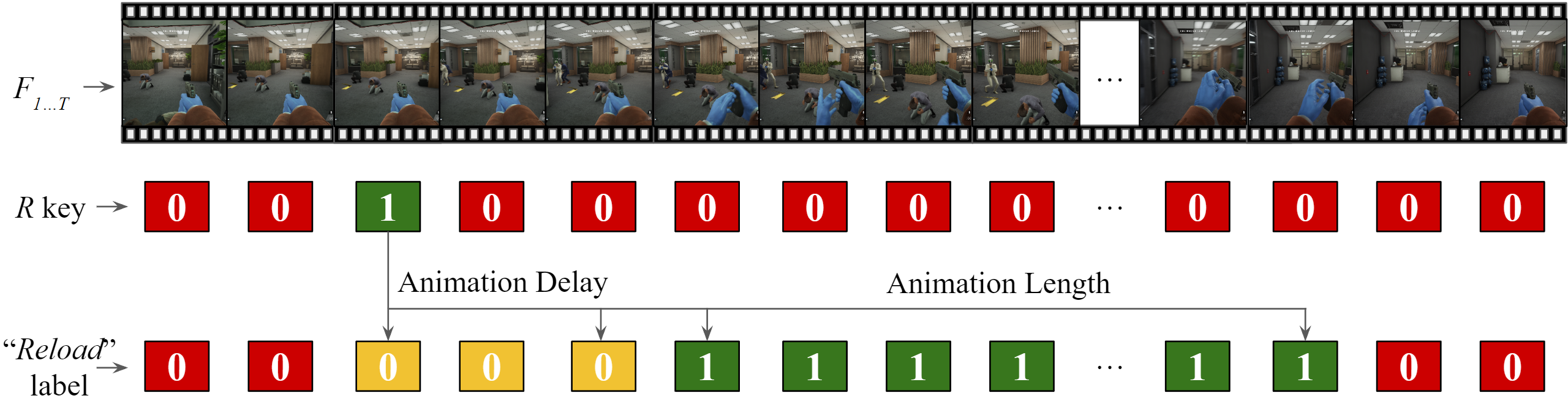}
    \caption{\textbf{Label Propagation} example for \textit{reload gun} animation activated by the player pressing the 'R' key. The yellow timesteps emphasise the delay in the onset of the animation visible on screen following the keypress, whereas the green timesteps indicate the complete duration during which the animation persists on screen. The animation \emph{delay} and \emph{length} hyper-parameters selected for each keypress are provided in Table \ref{tab:recordeddata}.}
  \label{fig:label_propagation}
\end{figure}

Not having access to game engines gives rise to another challenge. Each animation associated with different keypresses can exhibit different properties such as delayed onset of the said animation, as well as varying length/duration until the animation is completed on screen. This can lead to poor quality of synchronisation between animation frames and their respective action labels inferred via just raw keypresses. For instance, within the genre of FPS games, upon performing a keypress for actions such as \enquote{Move Forward} or \enquote{Shoot}, the associated animations are relatively more instantaneous and have a shorter duration compared to actions such as \enquote{Reload} or \enquote{Switch Item} which may last up to several seconds. Refer to the example in Figure \ref{fig:label_propagation} for a visual representation of this challenge.

\begin{figure}[!tb]
  \centering
   \includegraphics[clip,width=\columnwidth]{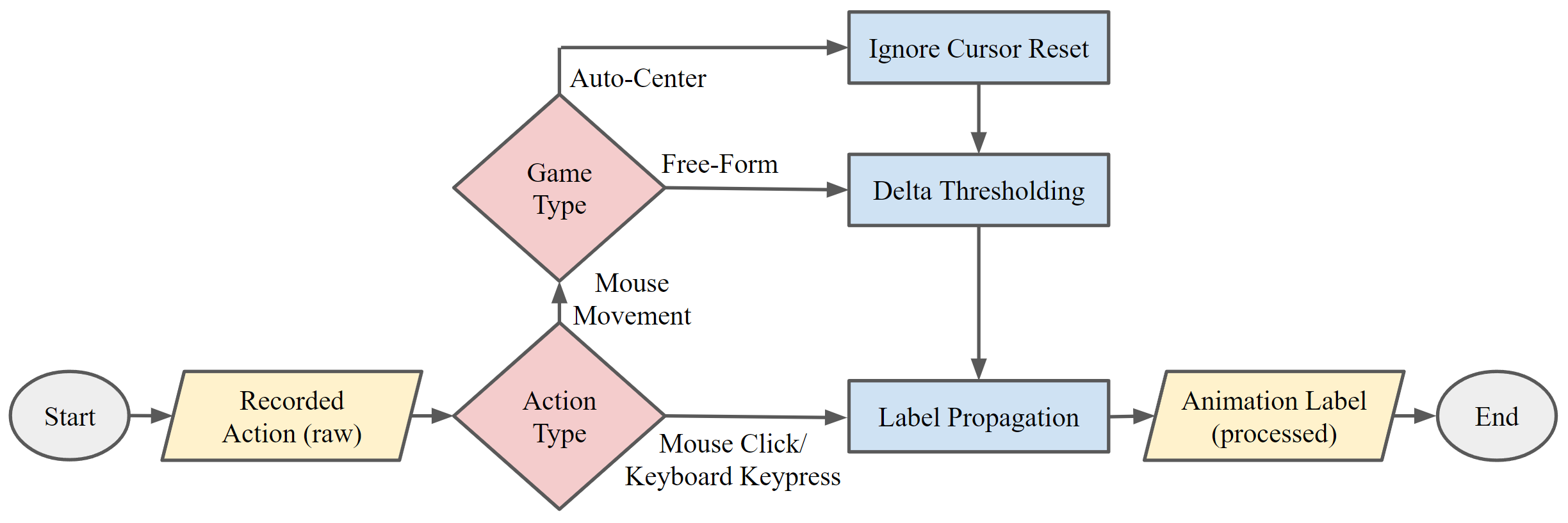}
    \caption{\textbf{Data Pre-Processing Pipeline} employed to transform keypress actions into animation labels in the SMG-25 dataset. Note that the additional datasets of CS:GO and Minecraft used in this study have been converted to a data format akin to SMG-25, and, subsequently, have been processed using this pipeline to ensure the uniformity and consistency of all data employed in our experiments.}
  \label{fig:preprocess_flowchart}
\end{figure}

\textbf{Label Propagation:} As a part of our data pre-processing step, we try to address the animation-related challenges with a simple technique we call \textit{label propagation}. This technique applies \textit{shift} and \textit{span} transformations to a keypress label propagating it forward along the temporal dimension. \textit{Shift} looks to account for animation \textit{delay} while \textit{span} tackles challenges related to animation \textit{length}. Figure \ref{fig:label_propagation} provides a visual example for this technique. This operation ensures that more frames within our dataset are associated with the correct animation label compared to using raw keypresses as a label. The \enquote{Animation} column of Table \ref{tab:recordeddata} provides the chosen hyper-parameters that indicate estimates of these animation-related properties (in terms of average number of frames), based on our observations across various FPS genre games. Figure \ref{fig:preprocess_flowchart} summarises all the steps of our data pre-processing pipeline.

\begin{figure}[!tb]
  \centering
  \begin{subfigure}[b]{\columnwidth}
    \includegraphics[clip,width=\columnwidth]{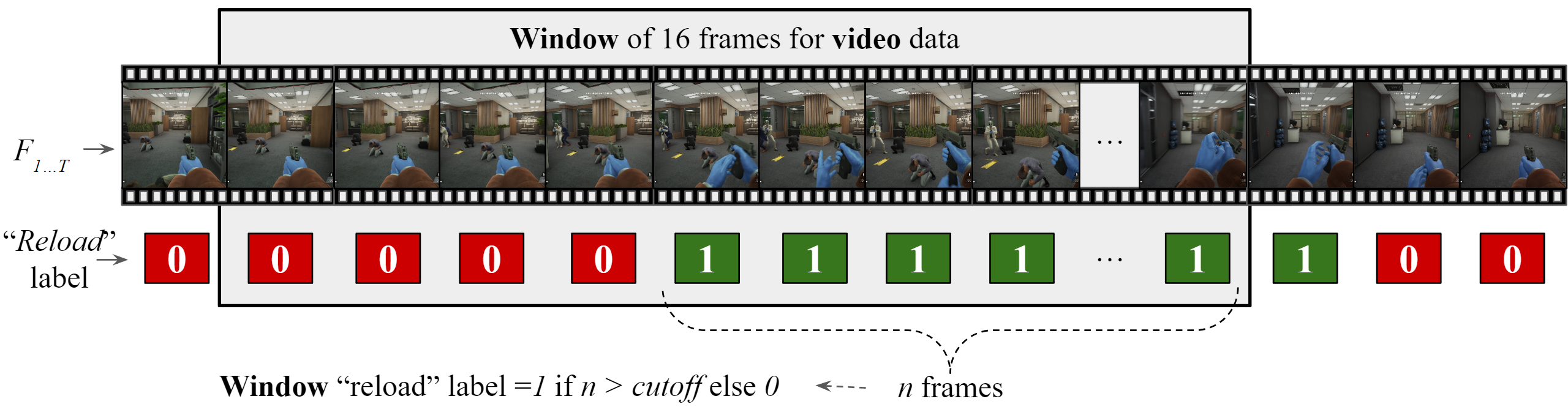}
    \caption{\textbf{Label Thresholding} example to ensure that sufficient length of animation is present within a window.}
  \label{subfig:label_thresholding}
  \end{subfigure}
  
  \vspace{10pt} 
  \begin{subfigure}[!tb]{\columnwidth}
    \centering
  \label{fig:textactionmap}
  
  \includegraphics[clip,width=\columnwidth]{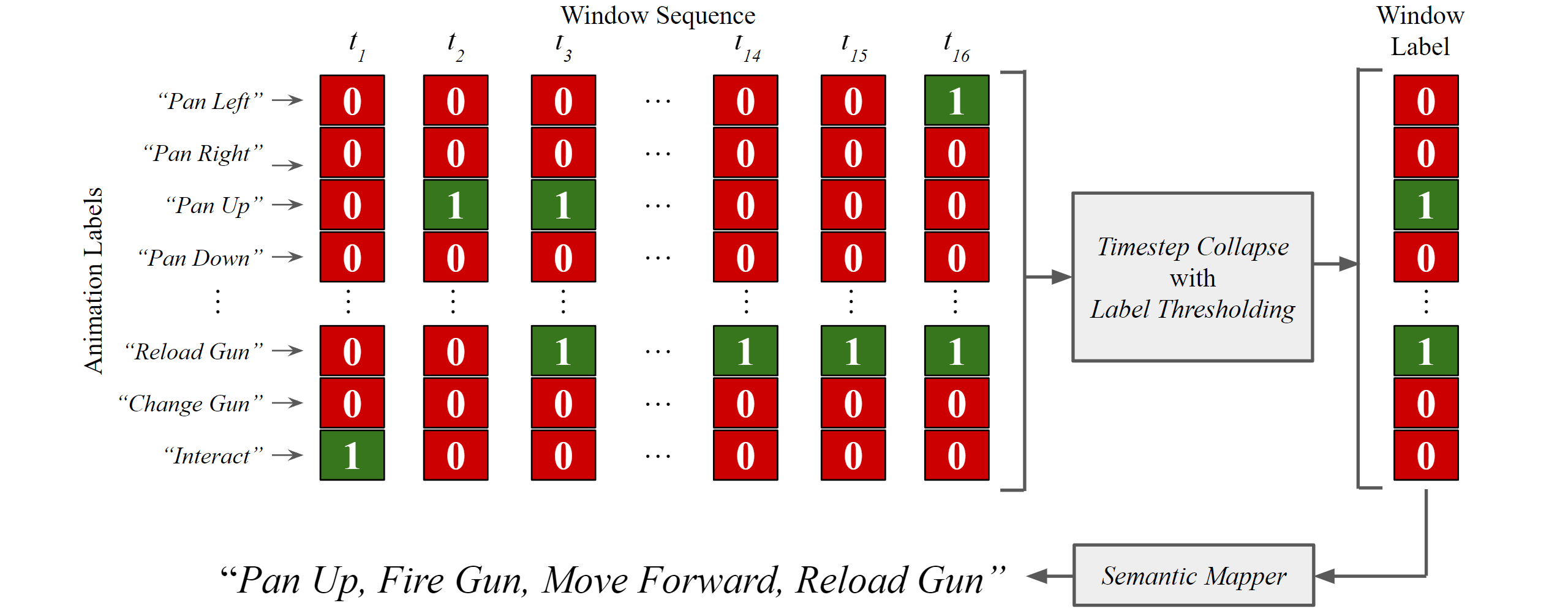}
  \caption{\textbf{Timestep Collapse} example for converting frame-wise animation labels to a window label in the form of a text caption.}
  \end{subfigure}
  \caption{Example showcasing how we combine frames and animation labels from consecutive timesteps to create associated text description of the actions performed in the time window of a video.}
  \label{subfig:timestep_collapse}
\end{figure}
\section{From Time-Steps to Time-Windows}

For the purpose of our research on player behaviour, as explained in Section 1, we opt to utilize videos instead of images as the input to our vision encoders. Consequently, we outline the necessary steps undertaken to transform the SMG-25 dataset, which includes synchronised data at the timestep level, into a format defined by time interval windows.

\subsection{Frames-to-Video Conversion}

The approach chosen involves stacking consecutive frames into a video sequence (as shown in Figure \ref{subfig:label_thresholding}), with each sequence representing a window size of 16 frames corresponding to approximately one second of gameplay. This selection is informed by pertinent studies on video backbone models \cite{wang2023videomaev2,wang2023masked} and methodologies specific to game analysis \cite{pearce2022counter,baker2022video}. A standard frame transformation involving resizing to dimensions $224 \times 224$ is applied, aligning with the specifications suitable for video representation models used in this study. Furthermore, to augment the dataset size during the frames-to-video conversion, a sliding window protocol with a stride of 8 (\emph{i.e.}, half the window size) is employed, resulting in twice the number of unique training samples as compared to using a non-overlapping sliding window protocol. Finally, we take measures to ensure that consecutive frames showcasing temporal discontinuity (\emph{e.g.}, due to any pause or resume of the gameplay; refer to Section \ref{appsubsec:smg25datacollection}), are not consolidated into a unified video sequence.

\subsection{Labels-to-Text Conversion}
\label{appsubsec:labelstotext}
Similar to combining consecutive frames into a video, we combine animation labels from each time-step to a single text description of the overall animation occurring in that respective time window. For this purpose we perform two sequential operations explained here.

\textbf{Timestep Collapse:} Illustrated in Figure \ref{subfig:timestep_collapse}, the timestep collapse operation is applied to sequences of animation labels within a given window, yielding a singular binary value for each animation label. This binary output indicates the presence or absence of the corresponding animations within the considered time window. Notably, we incorporate an intermediate \textit{Label Thresholding} operation, as depicted in Figure \ref{subfig:label_thresholding}. This additional step ensures that a sufficient amount of animation is visually perceptible within the window. A \textit{cutoff} hyper-parameter is introduced to establish a minimum threshold, determining whether a time window contains sufficient length of animation. The action-specific \textit{cutoff} values corresponding to each animation label are given in the  ``Animation'' column of Table \ref{tab:recordeddata}.

\textbf{Semantic Action Mapper:} This operation, illustrated in Figure \ref{subfig:timestep_collapse}, converts the singular binary values obtained from timestep collapse to a textual description of the animation occurring in the window. Each animation label is mapped to its relevant phrase describing the visual outcome of applying that animation within the game. We combine all obtained phrases for all observed animations with a \textit{comma} delimiter to form our overall text sequence for the window under consideration. This text sequence acts as our label for the associated video for the selected time window. Table \ref{tab:text_encodings} lists a number of phrasing alternatives that we experimented with in this study; we did not, however, observe any significant differences in the outcomes of any of our experiments. This suggests that the proposed behaviour alignment framework is robust to alternative phrasing as long as the phrasing is tight to the phrased action and the domain under investigation (\emph{i.e.}, FPS games in our study)

Collectively, these conversion steps contribute towards the creation of a dataset that contains synchronised pairs of videos and text description of all actions taken by the player within that video. As covered in Section 3.3 we use this dataset to align similar player behaviour across a diverse set of FPS games.

\begin{table}[t]
\centering
\caption{Lists of alternative phrase descriptions tested for each possible mouse action and keypress. The phrase description options are based on common terminology used within the genre of FPS games and are separated by ``$\mid$'' in the rightmost column of the table. All reported experimental results of this study rely on the phrases appearing in bold.}
% \vskip 0.15in
\resizebox{\textwidth}{!}{%
\begin{tabular}{c|c|l} \toprule
\textsc{\textbf{Device}} & \textsc{\textbf{Action/Key}} & \textsc{\textbf{Alternate Phrase Descriptions}} \\ \midrule
\multirow{6}{*}{\textsc{Mouse}} & \textsc{Move Left} & \textbf{\enquote{Pan Left}} \ $|$ \ \enquote{Rotate Left} \ $|$ \ \enquote{Look Left} \ $|$ \ \enquote{Turn Left} \\
& \textsc{Move Right} & \textbf{\enquote{Pan Right}} \ $|$ \ \enquote{Rotate Right} \ $|$ \ \enquote{Look Right} \ $|$ \ \enquote{Turn Right} \\
& \textsc{Move Up} & \textbf{\enquote{Pan Up}} \ $|$ \ \enquote{Rotate Up} \ $|$ \ \enquote{Look Up} \ $|$ \ \enquote{Turn Up} \\
& \textsc{Move Down} & \textbf{\enquote{Pan Down}} \ $|$ \ \enquote{Rotate Down} \ $|$ \ \enquote{Look Down} \ $|$ \ \enquote{Turn Down} \\
& \textsc{Left Click} & \textbf{\enquote{Fire Gun}} \ $|$ \ \enquote{Discharge Gun} \ $|$ \ \enquote{Attack} \ $|$ \ \enquote{Shoot Gun} \\
& \textsc{Right Click} & \textbf{\enquote{Aim Gun}} \ $|$ \ \enquote{Gun Scope} \\ \midrule
\multirow{10}{*}{\textsc{Keyboard}} & \textsc{W} & \textbf{\enquote{Move Forward}} \ $|$ \ \enquote{Advance} \ $|$ \ \enquote{Go Forward} \\
& \textsc{A} & \textbf{\enquote{Strafe Left}} \ $|$ \ \enquote{Move Left} \ $|$ \ \enquote{Go Left} \\
& \textsc{S} & \textbf{\enquote{Move Backward}} \ $|$ \ \enquote{Retreat} \ $|$ \ \enquote{Go Back} \\
& \textsc{D} & \textbf{\enquote{Strafe Right}} \ $|$ \ \enquote{Move Right} \ $|$ \ \enquote{Go Right} \\
& \textsc{R} & \textbf{\enquote{Reload Gun}} \ $|$ \ \enquote{Replace Clip} \\
& \textsc{Space} & \textbf{\enquote{Jump}} \ $|$ \ \enquote{Leap} \\
& \textsc{L. Shift} & \textbf{\enquote{Sprint}} \ $|$ \ \enquote{Run} \\
& \textsc{L. Ctrl, C} & \textbf{\enquote{Crouch}} \ $|$ \ \enquote{Squat} \\
& \textsc{1, 2, 3} & \textbf{\enquote{Change Gun}} \ $|$ \ \enquote{Switch Gun} \\
& \textsc{F} & \textbf{\enquote{Interact}} \\ \bottomrule
\end{tabular}}
% \vskip -0.1in
\label{tab:text_encodings}
\end{table}

\section{Behaviour Alignment}

Algorithm \ref{alg:behavepseudo} provides a PyTorch-like pseudocode implementation of the BehAVE framework depicted in Algorithm 1. Table \ref{apptab:alignmentscores} presents extended results from a comprehensive list of all configurations tested in the experiments reported in Section 5.1. Notably, in addition to varying the foundational video and text encoders, we also vary the alignment model architecture (see Section \ref{appsubsec:alignmentmodels}) as well as the loss function employed in the behaviour alignment training (see Section \ref{appsubsec:alignmentlosses}). 

\begin{algorithm}[t]
	\caption{BehAVE PyTorch-like pseudocode.} 
    \begin{algorithmic}
        \STATE \textcolor{Tan}{\texttt{\# semantic_{mapper}(.): convert action keypress to behaviour text caption}}
        \STATE \textcolor{Tan}{\texttt{\# video_{encoder}(.): frozen video foundation model}}
        \STATE \textcolor{Tan}{\texttt{\# text_{encoder}(.): frozen text foundation model}}
        \STATE \textcolor{Tan}{\texttt{\# alignment_projector(.): trainable alignment model}}
        \STATE 
        \STATE \texttt{for v, a in dataloader:
        \textcolor{Tan}{\# video sequence and corresponding actions}}
        \STATE \texttt{\qquad \textcolor{Tan}{\# convert keypresses to text sequence}}
        \STATE \texttt{\qquad a_c = semantic_{mapper}(a)}
        \STATE 
        \STATE \texttt{\qquad \textcolor{Tan}{\# obtain foundation encodings and L2 normalize}}
        \STATE \texttt{\qquad z_video = torch.normalize(video_{encoder}(v), p=2)}
        \STATE \texttt{\qquad z_caption = torch.normalize(text_{encoder}(a_c), p=2)}
        \STATE 
        \STATE \texttt{\qquad \textcolor{Tan}{\# train alignment model}}
        \STATE \texttt{\qquad z_align = torch.normalize(alignment_{projector}(z_video), p=2)}
        \STATE \texttt{\qquad loss = torch.cosine_embedding_loss(z_align, z_caption)}
        \STATE \texttt{\qquad loss.backward()}
        \STATE \texttt{\qquad update(alignment_{projector}.params) \textcolor{Tan}{\# adam optimizer}}
    \end{algorithmic}
    \label{alg:behavepseudo}
\end{algorithm}

\begin{table}[h]
  \centering
    \caption{Comparison of various alignment training configurations, including differences in pretrained video encoders, the number of layers in the projection model, action/text encoders, and the loss function. The resulting dimension of the shared space is denoted by the \enquote{Dim} column. Silhouette scores on the SMG-25 test set are presented, with average and standard deviation values calculated over 5 runs, each utilising a distinct, randomly selected set of games. Higher scores are preferable for behaviour-based cluster labels (indicated by $\uparrow$), while lower scores are optimal for game labels (indicated by $\downarrow$). The best-performing configuration in each category is highlighted in bold.}
\vskip 0.15in
\begin{sc}
\resizebox{\textwidth}{!}{%
  \begin{tabular}{lcccc|cccc}
    \toprule
    \multicolumn{5}{c|}{\textbf{Training Configuration}} & \multicolumn{3}{c}{\textbf{Behaviour Labels}} & \textbf{Game}  \\
    \textbf{Loss} & \textbf{Video Enc.} & \textbf{Proj.} & \textbf{Action Encoding} & \textbf{Dim.} & \textbf{Panning} $\uparrow$ & \textbf{Navigation} $\uparrow$ & \multicolumn{1}{c}{\textbf{Weapon} $\uparrow$} & \textbf{Label} $\downarrow$  \\ \midrule
     & VideoMAE &  &  & 768 & $0.08_{\pm0.00}$  & $0.13_{\pm0.00}$  & $0.03_{\pm0.00}$  & $0.10_{\pm0.00}$   \\
     - & MVD & - & - & 768 & $0.14_{\pm0.00}$ & $0.09_{\pm0.00}$ & $0.01_{\pm0.00}$ & $-0.05_{\pm0.00}$  \\ 
      & I3D &  &  & 512 & $0.03_{\pm0.00}$ & $0.04_{\pm0.00}$ & $0.00_{\pm0.00}$ & $0.07_{\pm0.00}$  \\ \midrule
    \multirow{4}{*}{Cosine} & \multirow{4}{*}{VideoMAE} & 1-MLP & \multirow{4}{*}{CLIP} & \multirow{4}{*}{512} & $0.27_{\pm0.00}$  & $0.35_{\pm0.00}$  & $0.15_{\pm0.00}$  & $-0.12_{\pm0.00}$   \\
     &  & 2-MLP & &  & $0.36_{\pm0.01}$  & $0.46_{\pm0.00}$  & $0.33_{\pm0.00}$  & $-0.21_{\pm0.01}$   \\
     &  & 3-MLP &  &  & $0.37_{\pm0.00}$  & $0.50_{\pm0.00}$  & $0.36_{\pm0.00}$  & $-0.22_{\pm0.01}$   \\
     &  & 4-MLP &  &  & $0.40_{\pm0.01}$  & $0.49_{\pm0.01}$  & $0.35_{\pm0.00}$  & $-0.20_{\pm0.01}$   \\ \midrule
    Pref. & \multirow{2}{*}{VideoMAE} & \multirow{2}{*}{4-MLP} & \multirow{2}{*}{CLIP} & \multirow{2}{*}{512} & $0.19_{\pm0.02}$  & $0.26_{\pm0.04}$  & $0.23_{\pm0.02}$  & $-0.14_{\pm0.01}$   \\
    MSE &  & &  & & $0.38_{\pm0.03}$  & $0.48_{\pm0.03}$  & $\mathbf{0.38_{\pm0.02}}$  & $-0.17_{\pm0.03}$ \\ 
    \midrule
    \multirow{3}{*}{Cosine} & VideoMAE & \multirow{3}{*}{4-MLP} & \multirow{3}{*}{Binary} & \multirow{3}{*}{16} & $0.35_{\pm0.00}$ & $0.48_{\pm0.01}$ & $0.32_{\pm0.01}$ & $-0.16_{\pm0.00}$ \\
     & MVD &  &  &  & $0.43_{\pm0.02}$ & $0.44_{\pm0.03}$ & $0.09_{\pm0.01}$  & $-0.24_{\pm0.02}$    \\
      & I3D & &  &  & $0.38_{\pm0.01}$ & $0.45_{\pm0.02}$ & $0.22_{\pm0.01}$ & $-0.21_{\pm0.01}$ \\
     \midrule
     \multirow{12}{*}{Cosine} & \multirow{12}{*}{VideoMAE} & \multirow{12}{*}{4-MLP} & AllMiniLM-L12-V2 \cite{reimers-2019-sentence-bert} & 384 & $0.41_{\pm0.03}$  & $0.48_{\pm0.04}$  & $0.35_{\pm0.02}$ & $-0.17_{\pm0.03}$  \\
     &  &  & AllMiniLM-L6-V2 \cite{reimers-2019-sentence-bert} & 384 & $0.42_{\pm0.02}$  & $0.49_{\pm0.05}$  & $0.31_{\pm0.01}$ & $-0.17_{\pm0.03}$  \\
     &  &  & Mobile BERT \cite{sun2020mobilebert} & 512 & $0.27_{\pm0.06}$  & $0.49_{\pm0.05}$  & $0.31_{\pm0.01}$ & $-0.17_{\pm0.03}$  \\
     &  &  & BERT \cite{devlin2018bert} & 768 & $0.48_{\pm0.06}$  & $\mathbf{0.60_{\pm0.04}}$  & $0.23_{\pm0.05}$ & $-0.24_{\pm0.07}$  \\
     &  &  & Flava \cite{singh2022flava} & 768 & $0.47_{\pm0.02}$  & $0.51_{\pm0.02}$  & $0.22_{\pm0.03}$ & $-0.24_{\pm0.05}$  \\
     &  &  & Data2Vec \cite{baevski2022data2vec} & 768 & $0.47_{\pm0.05}$  & $0.47_{\pm0.05}$  & $0.20_{\pm0.03}$ & $-0.23_{\pm0.04}$  \\
     &  &  & GPT \cite{radford2018improving} & 768 & $0.46_{\pm0.04}$  & $0.48_{\pm0.03}$  & $0.30_{\pm0.04}$ & $-0.18_{\pm0.05}$  \\
     &  &  & GPT-2 \cite{radford2019language} & 768 & $0.51_{\pm0.02}$  & $0.58_{\pm0.05}$  & $0.20_{\pm0.05}$ & $\mathbf{-0.29_{\pm0.06}}$  \\
     &  &  & All-MPNet-Base-V2 \cite{reimers-2019-sentence-bert} & 768 & $0.45_{\pm0.01}$  & $0.51_{\pm0.03}$  & $0.26_{\pm0.02}$ & $-0.22_{\pm0.03}$  \\
     &  &  & MPNet \cite{song2020mpnet} & 768 & $0.48_{\pm0.04}$  & $0.52_{\pm0.06}$  & $0.18_{\pm0.04}$ & $-0.25_{\pm0.06}$  \\
     &  &  & ALIGN \cite{jia2021scaling} & 768 & $\mathbf{0.52_{\pm0.03}}$  & $0.56_{\pm0.03}$ & $0.19_{\pm0.03}$  & $-0.26_{\pm0.04}$   \\ 
    &  &  & GPT Neo \cite{gpt-neo} & 2048 & $0.50_{\pm0.02}$  & $0.52_{\pm0.04}$  & $0.25_{\pm0.03}$ & $-0.21_{\pm0.03}$  \\ 
    \bottomrule
  \end{tabular}}
  \end{sc}
    \vskip -0.1in
  \label{apptab:alignmentscores}
\end{table}

\subsection{Varying the Size of the Alignment Model}
\label{appsubsec:alignmentmodels}

To thoroughly evaluate the alignment module of BehAVE we empirically alter the depth of the Multi-Layer Perceptron (MLP) head employed as our projection model for behaviour alignment training. Our experimental results, depicted in Table \ref{apptab:alignmentscores}, indicate performance enhancements when transitioning from a 1-layered network to a 4-layered network. Extending the depth beyond four layers, however, does not yield notable performance improvement, especially without a significant increase of the number of trainable parameters. The results corresponding to these experiments are reported in the \enquote{Proj.} column of Table \ref{apptab:alignmentscores} as $\#$-MLP, where $\#$ is the number of hidden layers. To enhance the generalisation capacity of BehAVE, dropout is applied between the linear layers at a rate of $40\%$, and Rectified Linear Unit (ReLU) activations are employed. All networks are trained using the \emph{adam} optimiser with a learning rate of $1e^{-4}$ and batches of size 128. The 4-layered MLP emerges as the optimal alignment projection model (with diminishing returns on increasing layers) for subsequent experiments, providing superior results while maintaining a reasonable number of training parameters. 

\subsection{Varying the Loss Function}
\label{appsubsec:alignmentlosses}

In addition to the cosine loss detailed in Section 3.3, we explore alternative loss functions, such as the mean-squared error (MSE) loss defined as follows:

\begin{equation}
\mathcal{L}_{\text{mse}}(z^{\text{align}}, z^{\text{caption}}) = \frac{1}{|z|} \sum_{n=1}^{|z|} (z^{\text{align}}_n - z^{\text{caption}}_n)^2
\end{equation}

The results reported on Table \ref{apptab:alignmentscores} indicate that alignment with MSE loss fails to achieve performance on par with the cosine loss. Consequently, we exclude this loss function from the remainder of our experiments. 

In contrast to the single-data-point focus of Cosine and MSE losses, we explore a pairwise comparison approach using the preference learning paradigm. This approach is motivated by the challenges discussed in Section \ref{appsubsec:dataprocessing}, where not every keypress corresponds to a visible animation, and vice versa. For example, a mouse \textit{left click} action may not lead to a gun fire animation within the game if the weapon is empty, hence keypresses are only considered as \enquote{proxy-labels} for annotating on-screen animations. To accommodate this nuance in our recorded dataset, we adopt a preference learning method that avoids strict assumptions about keypress labels. Thus, the \textit{conditional preference loss} in Equation \ref{eq:pref_loss} (with a margin hyper-parameter $\lambda$) can be conceptualised as a term that specifically considers the relative proximity of a video embedding to its corresponding text action embedding, when compared to the embedding of a different video. This formulation introduces a more lenient constraint on the alignment process, reducing susceptibility to training noise stemming from the \enquote{proxy-labels}.

\begin{dmath}
\mathcal{L}_{\text{pref}}(z^{\text{align}}_i, z^{\text{align}}_j \mid z^{\text{caption}}_i) = \max\left(0, \mathcal{L}_{\text{cos}}(z^{\text{align}}_i, z^{\text{caption}}_i) - \mathcal{L}_{\text{cos}}(z^{\text{align}}_j, z^{\text{caption}}_i) + \lambda\right)
\label{eq:pref_loss}
\end{dmath}

This loss focuses on the relative closeness between specific video-text pairs, providing a more robust training mechanism less sensitive to inaccuracies in the labelling process. However, as shown in Table \ref{apptab:alignmentscores}, alignment with such a preference loss did not yield significant benefits compared to the Cosine loss, leading us to discard this research paradigm for the current study.

\section{Behaviour Classification}
\label{app:behaviourclassification}
Section \ref{appsubsec:varyingvideoenc} provides results from additional video encoder configurations tested on the downstream task presented in Section 4.2, while Section \ref{appsubsec:idm} presents our preliminary efforts towards crafting a more fine-trained action prediction system instead of higher-level behaviour categories.

\subsection{Varying the Video Encoder}
\label{appsubsec:varyingvideoenc}
Table \ref{apptab:idm_behaviour_class} provides supplementary results pertaining to the behaviour classification transferability from CS:GO to unseen games from the SMG-25 test set. Notably, we illustrate the advantages of zero-shot transferability when employing alternative video encoders to VideoMAE, including the I3D or the MVD. Across all behaviour categories, we notice up to $22\%$ improvement on average for the I3D encoder and up to $17.5\%$ improvement on average for the MVD encoder. Neither of these alternate video encoders attain absolute performance levels comparable to VideoMAE. However, the comparative analysis between the representations of pretrained models and our proposed aligned embeddings underscores the robustness and effectiveness of our alignment framework regardless of the choice of the video encoder.

\begin{table}[!tb]
  \centering
  \caption{Transferability of behaviour classification analysed via classification accuracy for behaviour categories. The classifiers are trained on source game CS:GO and transferred to target domains of FPS games in SMG-25 test set. The transferability score is measured by percentage difference in accuracy between using aligned versus foundation encodings, averaged over 5 independent runs.}
  \vskip 0.15in
  \begin{sc}
    \resizebox{\textwidth}{!}{%
      \begin{tabular}{lc|ccc|ccc} 
        \toprule
        \multicolumn{1}{l}{\textbf{Encoder Methods}} & \multicolumn{1}{c|}{\textbf{Behaviour}} & \multicolumn{3}{c|}{\textbf{CS:GO (Source)}} & \multicolumn{3}{c}{\textbf{SMG-25 (Target)}} \\
        \multicolumn{1}{l}{\textbf{(Video - Action)}} & \multicolumn{1}{c|}{\textbf{Category}} & \textbf{Unaligned (\%)} & \textbf{Aligned (\%)} & \textbf{\%-Diff.} & \textbf{Unaligned (\%)} & \textbf{Aligned (\%)} & \textbf{\%-Diff.} \\ 
        \midrule
         & Panning & $83.32_{\pm0.49}$ & $75.93_{\pm2.24}$ & -8.88 & $67.22_{\pm1.74}$ & $72.04_{\pm2.57}$ & +7.29 \\
        VideoMAEv2 - CLIP & Navigation & $94.15_{\pm0.51}$ & $85.65_{\pm2.96}$ & -9.03 & $78.28_{\pm1.40}$ & $80.86_{\pm1.64}$ & +3.33 \\
         & Weapon & $92.68_{\pm0.66}$ & $86.65_{\pm0.41}$ & -6.51 & $66.14_{\pm2.11}$ & $80.86_{\pm0.69}$ & +22.34 \\
        \midrule 
         & Panning & $68.95_{\pm3.15}$ & $69.89_{\pm2.94}$ & +1.68 & $68.79_{\pm3.14}$ & $70.34_{\pm2.23}$ & +2.42 \\
        MVD - CLIP & Navigation & $63.86_{\pm6.87}$ & $62.36_{\pm2.64}$ & -1.32 & $53.73_{\pm2.92}$ & $62.91_{\pm5.89}$ & +17.55 \\
         & Weapon & $71.99_{\pm1.86}$ & $61.15_{\pm1.20}$ & -15.00 & $57.96_{\pm2.24}$ & $58.77_{\pm1.56}$ & +1.54 \\
        \midrule 
         & Panning & $79.94_{\pm1.36}$ & $72.11_{\pm1.75}$ & -9.53 & $60.30_{\pm3.90}$ & $68.61_{\pm1.29}$ & +14.25 \\
        I3D - CLIP & Navigation & $83.21_{\pm3.54}$ & $73.64_{\pm1.23}$ & -11.36 & $59.45_{\pm3.54}$ & $72.57_{\pm2.89}$ & +22.09 \\
         & Weapon & $87.79_{\pm0.90}$ & $72.98_{\pm2.26}$ & -16.87 & $62.78_{\pm2.66}$ & $68.26_{\pm2.06}$ & +8.80 \\
        \bottomrule
      \end{tabular}
    }
  \end{sc}
  \vskip -0.1in
  \label{apptab:idm_behaviour_class}
\end{table}

\begin{table}[!tb]
  \centering
  \caption{Transferability of behaviour classification from CS:GO (source) to SMG-25 test set (target). Actions having less than $30\%$ data points present in the training dataset are highlighted in gray.}
  \vskip 0.15in
  \begin{sc}
    \resizebox{\textwidth}{!}{%
      \begin{tabular}{lc|ccc|ccc} 
        \toprule
        \multicolumn{5}{c|}{\textbf{CS:GO (IDM Source)}} & \multicolumn{3}{c}{\textbf{SMG-25 (IDM Target)}} \\
        \multicolumn{1}{c}{\textbf{Action/Key}} & \multicolumn{1}{c}{\textbf{Freq. (\%)}} & \textbf{Unaligned (\%)} & \textbf{Aligned (\%)} & \textbf{\%-Diff.} & \textbf{Unaligned (\%)} & \textbf{Aligned (\%)} & \textbf{\%-Diff.} \\ 
        \midrule
        Move Forward & 69.83 & 90.26 & 85.01 & -5.82 & 81.91 & 84.27 & +2.88 \\
        Pan Right & 44.85 & 72.98 & 65.3 & -10.52 & 62.03 & 65.07 & +4.90 \\
        Strafe Right & 43.67 & 71.15 & 65.08 & -8.53 & 54.52 & 61.11 & +12.09 \\
        Pan Left & 43.07 & 73.55 & 64.24 & -12.66 & 58.23 & 65.83 & +13.05 \\
        Strafe Left & 41.37 & 75.09 & 67.11 & -10.63 & 55.32 & 60.76 & +9.83 \\
        Fire Gun & 30.19 & 94.36 & 89.51 & -5.14 & 77.53 & 88.09 & +13.62 \\
        \rowcolor{gray!30} % Add a grey background to the header row
        Pan Down & 14.10 & 76.8 & 72.94 & -5.03 & 58.66 & 56.12 & -4.33 \\
        \rowcolor{gray!30} % Add a grey background to this row
        Pan Up & 12.95 & 73.26 & 68.98 & -5.84 & 59.5 & 52.72 & -11.39 \\
        \rowcolor{gray!30} % Add a grey background to this row
        Move Backward & 10.82 & 69.16 & 62.32 & -9.89 & 55.47 & 66.79 & +20.41 \\
        \rowcolor{gray!30} % Add a grey background to this row
        Reload Gun & 4.77 & 92.86 & 58.83 & -36.65 & 51.91 & 53.75 & +3.54 \\
        \rowcolor{gray!30} % Add a grey background to this row
        Jump & 2.20 & 86.69 & 71.6 & -17.41 & 66.57 & 60.38 & -9.30 \\
        \rowcolor{gray!30} % Add a grey background to this row
        Change Gun & 1.58 & 100 & 100 & 0 & 60.32 & 55.56 & -7.89 \\
        \rowcolor{gray!30} % Add a grey background to this row
        Crouch & 0.78 & 81.08 & 71.62 & -11.67 & 51.8 & 46.4 & -10.42 \\
        \rowcolor{gray!30} % Add a grey background to the header row
        Aim Gun & 0.00 & - & - & - & - & - & - \\
        \rowcolor{gray!30} % Add a grey background to this row
        Sprint & 0.00 & - & - & - & - & - & - \\
        \rowcolor{gray!30} % Add a grey background to this row
        Interact & 0.00 & - & - & - & - & - & - \\ 
        \bottomrule
      \end{tabular}
    }
  \end{sc}
  \vskip -0.1in
  \label{apptab:idm_minecraft}
\end{table}

\subsection{From Behaviour Classification to Inverse Dynamics}
\label{appsubsec:idm}
In the context of video games, the actions executed by a player play a pivotal role in generating on-screen visuals, determined by the transition dynamics within the game world as defined by its underlying game engine. This section presents our preliminary efforts to extend the generalised FPS behaviour classification task towards the development of a more intricate generalised FPS Inverse Dynamics Model (IDM). In contrast to predicting overall player behaviour, our focus shifts towards predicting a specific set of actions enacted by the player, manifesting as raw keypresses.

It is noteworthy that the pretrained video backbone models employed herein are not explicitly designed to discern low-level patterns within video data. Their primary function, instead, is to identify high-level patterns, such as behaviour classes. Nonetheless, we endeavour to assess the extent of our capabilities by utilising frozen pretrained models without additional fine-tuning specific to our dataset or problem domain, aiming to recover precise actions through keypress predictions. This presents a more formidable challenge, as the objective involves identifying finer-grained actions as opposed to broader behaviour categories. Consequently, this section delineates two distinct approaches employed for a per-key IDM.

\textbf{Marginal Distribution of Actions:} The initial approach involves predicting each keypress through an independent classification head. In other words, the IDM outputs a marginal distribution for each action. Our model processes a single window and utilizes video input to predict the fine-grained label, denoting the presence or absence of each keypress. As this model forecasts primitive actions based on preceding and succeeding frames, it can be conceptualised as an inverse dynamics model. Table \ref{apptab:idm_minecraft} underscores the improved transferability capacity of BehAVE achieved in 8 out of 13 keypress actions. The outcomes, however, remain inconclusive for infrequently occurring keypresses, where data is limited (refer to ``Freq.'' column in Table \ref{apptab:idm_minecraft}). If we focus on actions characterised by an occurrence frequency over $30\%$, we observe improved transferability across all six keypress actions (see the white background rows of Table \ref{apptab:idm_minecraft}). The lack of adequate data for learning to discern certain keypresses and actions, coupled with resource constraints hindering fine-tuning of large video backbone models are the limiting factors for higher performance in less frequent actions. We can only expect that larger datasets containing representative samples for all actions (as in \cite{baker2022video}) and appropriate fine-tuning of models to the domain under consideration would yield IDMs of high accuracy and transferability capacity via BehAVE.

\textbf{Joint Distribution of Actions:} The second approach we employ involves the prediction of all possible actions within a given temporal window using an autoregressive transformer model. In essence, the IDM outputs a joint distribution of all potential actions. To implement this approach, we draw inspiration from the work of \cite{li2023decap} and leverage their methodology to decode our aligned video embeddings into textual captions that serve as a semantic representation of the on-screen animation resulting from all actions executed within the specified window. The objective is to predict these captions to match the window animation labels (as shown in Figure \ref{subfig:timestep_collapse}). To achieve this, we train a GPT 2 \cite{radford2019language} transformer to reconstruct the target window label, conditioned on the aligned latent. A notable difference in the text encoding strategy of our transformer involves treating each action label as a single, distinct token to ensure that a label for a single action remains atomic. This type of encoding ensures that our model perceives each individual keypress as a binary label, thereby eliminating the need to learn the vocabulary associated with our action text encoding. While we believe this approach has potential, we observe weak performance for this model likely due to insufficient training data for every possible action, and further investigation and improvements are an ongoing effort. 

Our preliminary findings demonstrate the potential of BehAVE to generate a transferable zero-shot IDM for FPS games; however, further data and computational effort is likely required to enhance its performance. Future research endeavours could focus on fine-tuning video backbone models rather than limiting the training to frozen models. Additionally, a prospective avenue for subsequent investigation could involve treating temporal windows as recurrent entities, as opposed to the independent and identically distributed predictions examined in the current study.

\bibliographystyle{splncs04}
\bibliography{main}